# Loss as the Inconsistency of a Probabilistic Dependency Graph: Choose Your Model, Not Your Loss Function

**Oliver E. Richardson**
Cornell University

## Abstract

In a world blessed with a great diversity of loss functions, we argue that that choice between them is not a matter of taste or pragmatics, but of model. Probabilistic dependency graphs (PDGs) are probabilistic models that come equipped with a measure of "inconsistency". We prove that many standard loss functions arise as the inconsistency of a natural PDG describing the appropriate scenario, and use the same approach to justify a well-known connection between regularizers and priors. We also show that the PDG inconsistency captures a large class of statistical divergences, and detail benefits of thinking of them in this way, including an intuitive visual language for deriving inequalities between them. In variational inference, we find that the ELBO, a somewhat opaque objective for latent variable models, and variants of it arise for free out of uncontroversial modeling assumptions—as do simple graphical proofs of their corresponding bounds. Finally, we observe that inconsistency becomes the log partition function (free energy) in the setting where PDGs are factor graphs.

## 1 INTRODUCTION

Many tasks in artificial intelligence have been fruitfully cast as optimization problems, but often the choice of objective is not unique. For instance, a key component of a machine learning system is a loss function which the system must minimize, and a wide variety of losses are used in pratice. Each implicitly represents different values and results in different behavior, so the choice between them can be quite important (Wang et al. 2020; Jadon 2020). Yet, because it's unclear how to choose a "good" loss function, the choice is usually made by empirics, tradition, and an instinctive calculus acquired through the practice—not by explicitly laying out beliefs. Furthermore, there is something to be gained by fiddling with these loss functions: one can add regularization terms, to (dis)incentivize (un)desirable behavior. But the process of tinkering with the objective until it works is often unsatisfying. It can be a tedious game without clear rules or meaning, while results so obtained are arguably overfitted and difficult to motivate.

By contrast, a choice of *model* admits more principled discussion, in part because models are testable; it makes sense to ask if a model is accurate. This observation motivates our proposal: instead of specifying a loss function directly, one articulates a situation that gives rise to it, in the (more interpretable) language of probablistic beliefs and certainties. Concretely, we use the machinery of Probabilistic Dependency Graphs (PDGs), a particularly expressive class of graphical models that can incorporate arbitrary (even inconsistent) probabilistic information in a natural way, and comes equipped with a well-motivated measure of inconsistency (Richardson and Halpern 2021).

A primary goal of this paper is to show that PDGs and their associated inconsistency measure can provide a "universal" model-based loss function. Towards this end, we show that many standard objective functions—cross entropy, square error, many statistical distances, the ELBO, regularizers, and the log partition function—arise naturally by measuring the inconsistency of the appropriate underlying PDG. This is somewhat surprising, since PDGs were not designed with the goal of capturing loss functions at all. Specifying a loss function indirectly like this is in some ways more restrictive, but it is also more intuitive (it no technical familiarity with losses, for instance), and admits more grounded defense and criticism.

For a particularly powerful demonstration, consider the variational autoencoder (VAE), an enormously successful class of generative model that has enabled breakthroughs in image generation, semantic interpolation, and unsupervised feature learning (Kingma and Welling 2014). Structurally, a VAE for a space $X$ consists of a (smaller) latent space $Z$, a prior distribution $p(Z)$, a decoder $d(X|Z)$, and an encoder $e(Z|X)$. A VAE is not considered a "graphical model" for two reasons. The first is that the encoder $e(Z|X)$ has the same target variable as $p(Z)$, so something like a Bayesian Network cannot simultaneously incorporate them both (besides, they could be inconsistent with one another). The second reason: it is not a VAE's structure, but rather its *loss function* that makes it tick. A VAE is typically trained by maximizing the "ELBO", a somewhat difficult-to-motivate function of a sample $x$, originating in variational calculus. We show that $-\text{ELBO}(x)$ is also precisely the inconsistency of a PDG containing $x$ and the probabilistic information of the autoencoder ($p, d$, and $e$). We can form such a PDG precisely because PDGs allow for inconsistency. Thus, PDG semantics simultaneously legitimize the strange structure of the VAE, and also justify its loss function, which can be thought of as a property of the model itself (its inconsistency), rather than some mysterious construction borrowed from physics.

Representing objectives as model inconsistencies, in addition to providing a principled way of selecting an objective, also has beneficial pedagogical side effects, because of the *structural* relationships between the underlying models. For instance, these relationships will allow us to derive simple and intuitive visual proofs of technical results, such as the variational inequalities that traditionally motivate the ELBO, and the monotonicity of Rényi divergence.

In the coming sections, we show in more detail how this concept of inconsistency, beyond simply providing a permissive and intuitive modeling framework, reduces exactly to many standard objectives used in machine learning and to measures of statistical distance. We demonstrate that this framework clarifies the relationships between them, by providing clear derivations of otherwise opaque inequalities.

## 2 PRELIMINARIES

We generally use capital letters for variables, and lower case letters for their values. For variables $X$ and $Y$, a conditional probability distribution (cpd) $p$ on $Y$ given $X$, written $p(Y|X)$, consists of a probability distribution on $Y$ (denoted $p(Y|X = x)$ or $p(Y|\,x)$ for short), for each possible value $x$ of $X$. If $\mu$ is a probability on outcomes that determine $X$ and $Y$, then $\mu(X)$ denotes the marginal of $\mu$ on $X$, and $\mu(Y|X)$ denotes the conditional marginal of $\mu$ on $Y$ given $X$. Depending on which we find clearer in context, we write either $\mathbb{E}_\mu f$ or $\mathbb{E}_{\omega\sim\mu} f(\omega)$ for expectation of $f : \Omega \to \mathbb{R}$ over a distribution $\mu$ with outcomes $\Omega$. We write $\boldsymbol{D}(\mu \parallel \nu) = \mathbb{E}_\mu \log \frac{\mu}{\nu}$ for the relative entropy (KL Divergence) of $\nu$ with respect to $\mu$, we write $\mathrm{H}(\mu) := \mathbb{E}_\mu \log \frac{1}{\mu}$ for the entropy of $\mu$, $\mathrm{H}_\mu(X) := \mathrm{H}(\mu(X))$ for the marginal entropy on a variable $X$, and $\mathrm{H}_\mu(Y \mid X) := \mathbb{E}_\mu \log {}^1\!/\mu(Y|X)$ for the conditional entropy of $Y$ given $X$.

A *probabilistic dependency graph* (PDG) (Richardson and Halpern 2021), like a Bayesian Network (BN), is a directed graph with cpds attached to it. While this data is attached to the *nodes* of a BN, it is attached to the *edges* of a PDG. For instance, a BN of shape $X \to Y \leftarrow Z$ contains a single cpd $\Pr(Y|X,Z)$ on $Y$ given joint values of $X$ and $Z$, while a PDG of the same shape contains two cpds $p(Y|X)$ and $q(Y|Z)$. The second approach is strictly more expressive, and can encode joint dependence with an extra variable. All information in a PDG can be expressed with variable confidence. We now restate the formal definition.

**Definition 1.** A Probabilistic Dependency Graph (PDG) is a tuple $\mathcal{M} = (\mathcal{N}, \mathcal{E}, \mathcal{V}, \mathbf{p}, \boldsymbol{\alpha}, \boldsymbol{\beta})$, where

- $\mathcal{N}$ is a set of nodes, corresponding to variables;
- $\mathcal{V}$ associates each node $X \in \mathcal{N}$ with a set $\mathcal{V}(X)$ of possible values that the variable $X$ can take;
- $\mathcal{E}$ is a set of labeled edges $\{X \xrightarrow{L} Y\}$, each with a source $X$ and target $Y$ from $\mathcal{N}$;
- $\mathbf{p}$ associates a cpd $\mathbf{p}_L(Y|X)$ to each edge $X \xrightarrow{L} Y \in \mathcal{E}$;
- $\boldsymbol{\alpha}$ associates to each edge $X \xrightarrow{L} Y$ a non-negative number $\alpha_L$ representing the modeler's confidence in the functional dependence of $Y$ on $X$;
- $\boldsymbol{\beta}$ associates to each edge $L$ a number $\beta_L$, the modeler's confidence in the reliability of the cpd $\mathbf{p}_L$. □

How should one choose parameters $\boldsymbol{\beta}$ and $\boldsymbol{\alpha}$? A choice of $\beta_L = 0$ means that the cpd $p_L$ is effectively ignored, in the sense that such a PDG is equivalent to one in which the edge is attached to a different cpd $q \neq p_L$. On the other hand, a large value of $\beta_L$ (or $\infty$) indicates high (or absolute) confidence in the cpd. By default, we suppose $\beta = 1$, which is just a convenient choice of units—what's important are the magnitudes of $\beta$ relative to one another. The parameter $\boldsymbol{\alpha}$, typically in $[0, 1]$, represents certainty in the causal structure of the graph, and plays only a minor role in this paper.

Like other graphical models, PDGs have semantics in terms of joint distributions $\mu$ over all variables. Most directly, a PDG $\mathcal{M}$ determines two scoring functions on joint distributions $\mu$. For the purposes of this paper, the more important of the two is the *incompatibility* of $\mu$ with respect to $\mathcal{M}$, which measures the quantitative discrepency between $\mu$ and $\mathcal{M}$'s cpds, and is given by

$$Inc_{\mathcal{M}}(\mu) := \sum_{X \xrightarrow{L} Y} \beta_L \mathop{\mathbb{E}}_{x \sim \mu(X)} \boldsymbol{D}\Big(\mu(Y|x) \,\Big\|\, \mathbf{p}_L(Y|x)\Big). \quad (1)$$

Relative entropy $\boldsymbol{D}(\mu\|p)$ measures divergence between $\mu$ and $p$, and can be viewed as the overhead (in extra bits per sample) of using codes optimized for $p$, when in fact samples are distributed according to $\mu$ (MacKay 2003). But if one uses edges in proportion to the confidence one has in them, then inefficiencies for of high-confidence cpds are compounded, and hence more costly. So $Inc_{\mathcal{M}}(\mu)$ measures the total excess cost of using $\mathcal{M}$'s cpds in proportion to their confidences $\boldsymbol{\beta}$, when worlds are distributed according to $\mu$.

The *inconsistency* of $\mathcal{M}$, denoted $\langle\!\langle \mathcal{M} \rangle\!\rangle$, is the smallest possible incompatibility of $\mathcal{M}$ with any distribution: $\langle\!\langle \mathcal{M} \rangle\!\rangle := \inf_\mu Inc_{\mathcal{M}}(\mu)$. This quantity, which does not depend on $\boldsymbol{\alpha}$, is the primary focus of this paper.

The second scoring function defined by a PDG $\mathcal{M}$, called the *I*nformation *Def*iciency, measures the qualitative discrepency between $\mathcal{M}$ and $\mu$, and is given by

$$IDef_{\mathcal{M}}(\mu) := -\mathrm{H}(\mu) + \sum_{X \xrightarrow{L} Y} \alpha_L \mathrm{H}_\mu(Y \mid X).$$

$IDef_{\mathcal{M}}(\mu)$ can be thought of as the information needed to separately describe the target of each edge $L$ given the value of its source (weighted by $\alpha_L$) beyond the information needed to fully describe a sample from $\mu$.

As shown by Richardson and Halpern (2021), it is via these two scoring functions that PDGs capture other graphical models. The distribution specified by a BN $\mathcal{B}$ is the unique one that minimizes both $Inc_{\mathcal{B}}$ and $IDef_{\mathcal{B}}$ (and hence every positive linear combination of the two), while the distribution specfied by a factor graph $\Phi$ uniquely minimizes the sum $Inc_\Phi + IDef_\Phi$. In general, for any $\gamma > 0$, one can consider a weighted combination $[\![\mathcal{M}]\!]_\gamma(\mu) := Inc_{\mathcal{M}}(\mu) + \gamma \; IDef_{\mathcal{M}}(\mu)$, for which there is a corresponding $\gamma$-inconsistency $\langle\!\langle \mathcal{M} \rangle\!\rangle_\gamma := \inf_\mu [\![\mathcal{M}]\!]_\gamma(\mu)$. In the limit as $\gamma \to 0$, there is always a unique best distribution whose score is $\langle\!\langle \mathcal{M} \rangle\!\rangle$.

We now present some shorthand to clarify the presentation. We typically conflate a cpd's symbol with its edge label, thus drawing the PDG with a single edge attached to $f(Y|X)$ as $\boxed{X} \xrightarrow{f} \boxed{Y}$. Definition 1 is equivalent to one in which edge sources and targets are both *sets* of variables. This allows us to indicate joint dependence with multi-tailed arrows, joint distributions with multi-headed arrows, and unconditional distributions with nothing at the tail. For instance, we draw

$p(Y|X,Z)$ as 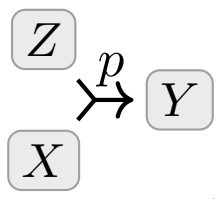, and $q(A,B)$ as 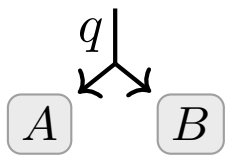.

To emphasize that a cpd $f(Y|X)$ is degenerate (a function $f : X \to Y$), we will draw it with two heads, as in: $\boxed{X} \overset{f}{\twoheadrightarrow} \boxed{Y}$. We identify an event $X{=}x$ with the degenerate unconditional distribution $\delta_x(X)$ that places all mass on $x$; hence it may be associated to an edge and drawn simply as $\overset{x}{\twoheadrightarrow} \boxed{X}$. To specify a confidence $\beta \neq 1$, we place the value near the edge, lightly colored and parenthesized, as in: $\xrightarrow[(\beta)]{p} \boxed{X}$, and we write $(\infty)$ for the limit of high confidence ($\beta \to \infty$).

Intuitively, believing more things can't make you any less inconsistent. Lemma 1 captures this formally: adding cpds or increasing confidences cannot decrease a PDG's inconsistency.

**Lemma 1** (Monotonicity of $\langle\!\langle \cdot \rangle\!\rangle$)**.** *Suppose PDGs $\mathcal{M}$ and $\mathcal{M}'$ differ only in their edges (resp. $\mathcal{E}$ and $\mathcal{E}'$) and confidences (resp. $\boldsymbol{\beta}$ and $\boldsymbol{\beta}'$). If $\mathcal{E} \subseteq \mathcal{E}'$ and $\beta_L \leq \beta'_L$ for all $L \in \mathcal{E}$, then $\langle\!\langle \mathcal{M} \rangle\!\rangle_\gamma \leq \langle\!\langle \mathcal{M}' \rangle\!\rangle_\gamma$ for all $\gamma$.*[1] [link to proof]

As we will see, this tool is sufficient to derive many interesting relationships between loss functions.

## 3 STANDARD METRICS AS INCONSISTENCIES

Suppose you believe that $X$ is distributed according to $p(X)$, and also that it (certainly) equals some value $x$. These beliefs are consistent if $p(X{=}x) = 1$ but become less so as $p(X = x)$ decreases. In fact, this inconsistency is equal to the information content $\mathrm{I}_p[X{=}x] := -\log p(X{=}x)$, or *surprisal* (Tribus 1961), of the event $X{=}x$, according to $p$.[2] In machine learning, $\mathrm{I}_p$ is usually called "negative log likelihood", and is perhaps the most popular objective for training generative models (Grover and Ermon 2018; Myung 2003).

**Proposition 2.** *Consider a distribution $p(X)$. The inconsistency of the PDG comprising $p$ and $X = x$ equals the surprisal $\mathrm{I}_p[X{=}x]$. That is,* [link to proof]

$$\mathrm{I}_p[X{=}x] = \left\langle\!\!\left\langle \xrightarrow{p} \boxed{X} \overset{x}{\twoheadleftarrow} \right\rangle\!\!\right\rangle.$$

*(Recall that $\langle\!\langle \mathcal{M} \rangle\!\rangle$ is the inconsistency of the PDG $\mathcal{M}$.)*

In some ways, this result is entirely unsurprising, given that (1) is a flexible formula built out of information theoretic primitives. Even so, note that the inconsistency of believing both a distribution and an event happens to be the standard measure of discrepency between the two—and is even named after "surprise", a particular expression of epistemic conflict.

Still, we have a ways to go before this amounts to any more than a curiosity. One concern is that this picture is incomplete; we train probabilistic models with more than one sample. What if we replace $x$ with an empirical distribution over many samples?

[1] All proofs can be found in Appendix C.

[2] This construction requires the event $X{=}x$ to be measurable. One can get similar, but subtler, results for densities, where this is not the case; see Appendix A.

[link to proof]

**Proposition 3.** *If $p(X)$ is a probabilistic model of $X$, and $\mathcal{D} = \{x_i\}_{i=1}^m$ is a dataset with empirical distribution $\Pr_{\mathcal{D}}$, then* $\quad \mathrm{CrossEntropy}(\Pr_{\mathcal{D}}, p) =$

$$\frac{1}{m}\sum_{i=1}^m \mathrm{I}_p[X{=}x_i] = \left\langle\!\!\left\langle \xrightarrow{p} \boxed{X} \xleftarrow[(\infty)]{\Pr_{\mathcal{D}}} \right\rangle\!\!\right\rangle + \mathrm{H}(\Pr_{\mathcal{D}}).$$

**Remark 1.** *The term $H(\Pr_{\mathcal{D}})$ is a constant depending only on the data, so is irrelevant for optimizing $p$.*

Essentially the only choices we've made in specifying the PDG of Proposition 3 are the confidences. But $\mathrm{CrossEntropy}(\Pr_{\mathcal{D}}, p)$ is the expected code length per sample from $\Pr_{\mathcal{D}}$, when using codes optimized for the (incorrect) distribution $p$. So implicitly, a modeler using cross-entropy has already articulated a belief the data distribution $\Pr_{\mathcal{D}}$ is the "true one". To get the same effect from a PDG, the modeler must make this belief explicit by placing infinite confidence in $\Pr_{\mathcal{D}}$.

Now consider an orthogonal generalization of Proposition 2, in which the sample $x$ is only a partial observation of $(x, z)$ from a joint model $p(X, Z)$.

[link to proof]

**Proposition 4.** *If $p(X, Z)$ is a joint distribution, then the information content of the partial observation $X = x$ is given by*

$\mathrm{I}_p[X{=}x] =$ 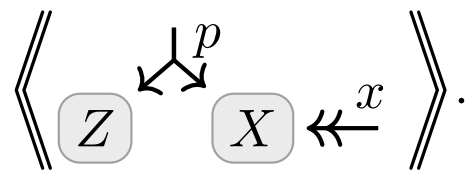 . (2)

Intuitively, the inconsistency of the PDG on the right side of (2) is localized to $X$, where the observation $x$ conflicts with $p(X)$; other variables don't make a difference. The multi-sample partial-observation generalization also holds; see Appendix B.3.

So far we have considered models of an unconditional distribution $p(X)$. Because they are unconditional, such models must describe how to generate a complete sample $X$ without input, and so are called *generative*; the process of training them is called *unsupervised* learning (Hastie, Tibshirani, and Friedman 2009). In the (more common) *supervised* setting, we train *discriminative* models to predict $Y$ from $X$, via labeled samples $\{(x_i, y_i)\}_i$. There, cross entropy loss is perhaps even more dominant—and it is essentially the inconsistency of a PDG consisting of the predictor $h(Y|X)$ together with high-confidence data.

[link to proof]

**Proposition 5** (Cross Entropy, Supervised)**.** *The inconsistency of the PDG comprising a probabilistic predictor $h(Y|X)$, and a high-confidence empirical distribution $\Pr_{\mathcal{D}}$ of a dataset $\mathcal{D} = \{(x_i, y_i)\}_{i=1}^m$ equals the cross-entropy loss (minus the empirical uncertainty in $Y$ given $X$, a constant depending only on $\mathcal{D}$). That is,*

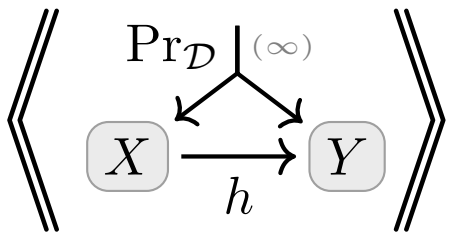 $= \frac{1}{m}\sum_{i=1}^m \log \frac{1}{h(y_i \mid x_i)} - \mathrm{H}_{\Pr_{\mathcal{D}}}(Y|X).$

Simple evaluation metrics, such as the accuracy of a classifier, and the mean squared error of a regressor, also arise naturally as inconsistencies.

[link to proof]

**Proposition 6** (Log Accuracy as Inconsistency)**.** *Consider functions $f, h : X \to Y$ from inputs to labels, where $h$ is a predictor and $f$ generates the true labels. The inconsistency of believing $f$ and $h$ (with any confidences), and a distribution $D(X)$ with confidence $\beta$, is $\beta$ times the log accuracy of $h$. That is,*

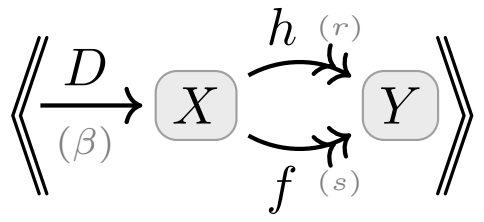 $= -\beta \log \Pr_{x\sim D}(f(x){=}h(x)) = \beta\, \mathrm{I}_D[f = h].$ (3)

One often speaks of the accuracy of a hypothesis $h$, leaving the true labels $f$ and empirical distribution $D$ implicit. Yet Proposition 6 suggests that there is a sense in which $D(X)$ plays the primary role: the inconsistency in (3) is scaled by the confidence in $D$, and does not depend on the confidences in $h$ or $f$. Why should this be this the case? Expressing $(x, y)$ such that $y \neq f(x)$ with codes optimized for $f$ is not just inefficient, but impossible. The same is true for $h$, so we can only consider $\mu$ such that $\mu(f = h) = 1$. In other words, the only way to form a joint distribution *at all* compatible with both the predictor $h$ and the labels $f$, is to throw out samples that the predictor gets wrong—and the cost of throwing out samples scales with your confidence in $D$, not in $h$. This illustrates why accuracy gives no gradient information for training $h$. It is worth noting that this is precisely the opposite of what happened in Proposition 5: there we were unwilling to budge on the input distribution, and the inconsistency scaled with the confidence in $h$.

Observe how even properties of these simple metrics—relationships with one another and features of gradients—can be clarified by an underlying model.

When $Y \cong \mathbb{R}^n$, an estimator $h(Y|X)$ is referred to as a regressor instead of a classifier. In this setting, most answers are incorrect, but some more so than others. A common way of measuring incorrectness is with mean squared error (MSE): $\mathbb{E}\,|f(X) - Y|^2$. MSE is also the inconsistency of believing that the labels and predictor have Gaussian noise—often a reasonable assumption because of the central limit theorem.

[link to proof]

**Proposition 7** (MSE as Inconsistency)**.**

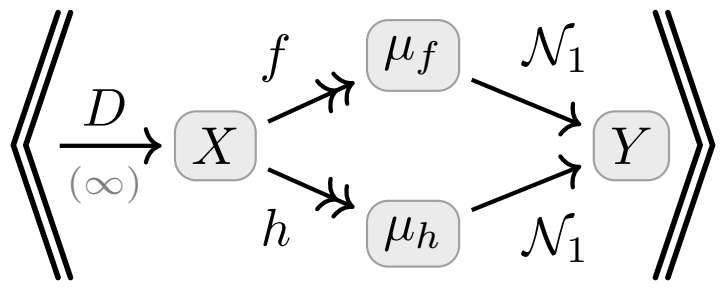 $= \frac{1}{2}\,\mathbb{E}_D\big|f(X) - h(X)\big|^2 =: \mathrm{MSE}_D(f, h)\,,$

*where $\mathcal{N}_1(Y \mid \mu)$ is a unit Gaussian on $Y$ with mean $\mu$.*

In the appendix, we treat general univariate Gaussian predictors, with arbitrary variances and confidences.

## 4 REGULARIZERS AND PRIORS

Regularizers are extra terms added to loss funtions, which provide a source of inductive bias towards simple model parameters. There is a well-known correspondence between using a regularizer and doing maximum *a posteriori* inference with a prior,[3] in which L2 regularization corresponds to a Gaussian prior (Rennie 2003), while L1 regularization corresponds to a Laplacian prior (Williams 1995). Note that the ability to make principled modeling choices about regularizers is a primary benefit of this correspondence. Our approach provides a new justification of it.

link to proof

**Proposition 8.** *Suppose you have a parameterized model $p(Y|\Theta)$, a prior $q(\Theta)$, and a trusted distribution $D(Y)$. The inconsistency of also believing $\Theta = \theta$ is the cross entropy loss, plus the regularizer: $\log \frac{1}{q(\theta)}$ times your confidence in $q$. That is,*

$$\left\langle\!\!\left\langle \xrightarrow[(\beta)]{q} \boxed{\Theta} \xrightarrow{p} \boxed{Y},\ \xrightarrow{\theta} \boxed{\Theta},\ \xrightarrow[(\infty)]{D} \boxed{Y} \right\rangle\!\!\right\rangle = \mathop{\mathbb{E}}_{y\sim D} \log \frac{1}{p(y\,|\,\theta)} + \beta \log \frac{1}{q(\theta)} - \mathrm{H}(D) \tag{4}$$

If our prior is $q(\theta) = \frac{1}{k}\exp(-\frac{1}{2}\theta^2)$, a (discretized) unit gaussian, then the right hand side of (4) becomes

$$\underbrace{\mathbb{E}_D \log \frac{1}{p(Y\,|\,\theta)}}_{\substack{\text{Cross entropy loss} \\ \text{(data-fit cost of } \theta)}} + \underbrace{\frac{\beta}{2}\theta_0}_{\substack{\text{L2 regularizer} \\ \text{(complexity cost of } \theta)}} + \underbrace{\beta \log k - \mathrm{H}(D)}_{\text{constant in } p \text{ and } \theta},$$

which is the L2 regularized version of Proposition 3. Moreover, the regularization strength corresponds exactly to the confidence $\beta$. What about other priors? It is not difficult to see that if we use a (discretized) unit Laplacian prior, $q(\theta) \propto \exp(-|\theta|)$, the second term instead becomes $\beta|\theta_0|$, which is L1 regularization. More generally, to consider a complexity measure $U(\theta)$, we need only include the Gibbs distribution $\Pr_U(\theta) \propto \exp(-U(\theta))$ into our PDG. We remark that nothing here is specific to cross entropy; any of the objectives we describe can be regularized in this way.

## 5 STATISTICAL DISTANCES AS INCONSISTENCIES

Suppose you are concerned with a single variable $X$. One friend has told you that it is distributed according to $p(X)$; another has told you that it follows $q(X)$. You adopt both beliefs. Your mental state will be inconsistent if (and only if) $p \neq q$, with more inconsistency the more $p$ and $q$ differ. Thus the inconsistency of a PDG comprising $p$ and $q$ is a measure of divergence. Recall that a PDG also allows us to specify the confidences $\beta_p$ and $\beta_q$ of each cpd, so we can form a PDG divergence $\boldsymbol{D}^{\mathrm{PDG}}_{(r,s)}(p\|q)$ for every setting $(r,s)$ of $(\beta_p, \beta_q)$. It turns out that a large class of statistical divergences arise in this way. We start with a familiar one.

**Proposition 9** (KL Divergence as Inconsistency)**.** *The inconsistency of believing $p$ with complete certainty, and also $q$ with some finite certainty $\beta$, is $\beta$ times the KL Divergence (or relative entropy) of $q$ with respect to $p$. That is,*

$$\left\langle\!\!\left\langle \xrightarrow[(\infty)]{p} \boxed{X} \xleftarrow[(\beta)]{q} \right\rangle\!\!\right\rangle = \beta\, \boldsymbol{D}(p \parallel q).$$

This result gives us an intuitive interpretation of the asymmetry of relative entropy / KL divergence, and a prescription about when it makes sense to use it. $\boldsymbol{D}(p \parallel q)$ is the inconsistency of a mental state containing both $p$ and $q$, when absolutely certain of $p$ (and not willing to budge on it). This concords with the standard intuition that $\boldsymbol{D}(p \parallel q)$ reflects the amount of information required to change $q$ into $p$, which is why it is usually called the relative entropy "from $q$ to $p$".

We now consider the general case of a PDG comprising $p(X)$ and $q(X)$ with arbitrary confidences.

**Lemma 10.** *The inconsistency $\boldsymbol{D}^{\mathrm{PDG}}_{(r,s)}(p\|q)$ of a PDG comprising $p(X)$ with confidence $r$ and $q(X)$ with confidence $s$ is given in closed form by* link to proof

$$\left\langle\!\!\left\langle \xrightarrow[(r)]{p} \boxed{X} \xleftarrow[(s)]{q} \right\rangle\!\!\right\rangle = -(r+s) \log \sum_x \Big( p(x)^r q(x)^s \Big)^{\frac{1}{r+s}}.$$

Of the many generalizations of KL divergence, Rényi divergences, first characterized by Alfréd Rényi 1961 are perhaps the most significant, as few others have found either application or an interpretation in terms of coding theory (Van Erven and Harremos 2014). The Rényi divergence of order $\alpha$ between two distributions $p(X)$ and $q(X)$ is given by

$$\boldsymbol{D}_\alpha(p \parallel q) := \frac{1}{1-\alpha} \log \sum_{x \in \mathcal{V}(X)} p(x)^\alpha q(x)^{1-\alpha}. \tag{5}$$

Rényi introduced this measure in the same paper as the more general class of $f$-divergences, but directs his attention towards those of the form (5), because they satisfy a natural weakening of standard postulates for Shannon entropy due to Fadeev (1957). Concretely, every symmetric, continuous measure that additively separates over independent events, and with a certain "mean-value property", up to scaling, is of the form (5) for some $\alpha$ (Rényi 1961). It follows from Lemma 10 that every Rényi divergence is a PDG divergence, and every (non-limiting) PDG divergence is a (scaled) Rényi divergence.

[3] A full account can be found in the appendix.

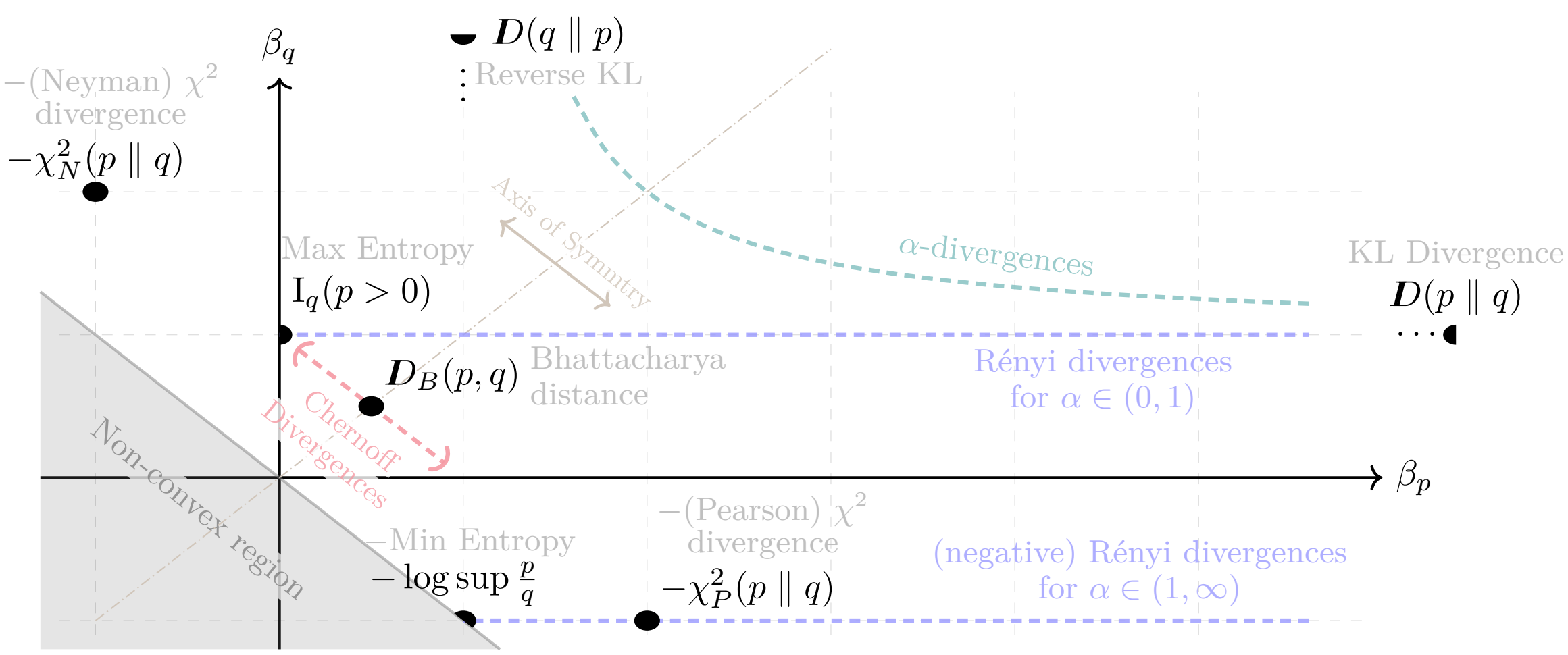

Figure 1: A map of the inconsistency of the PDG comprising $p(X)$ and $q(X)$, as we vary their respective confidences $\beta_p$ and $\beta_q$. Solid circles indicate well-known named measures, semicircles indicate limiting values, and the heavily dashed lines are well-established classes.

**Corollary 10.1** (Rényi Divergences)**.**

$$\left\langle\!\!\left\langle \xrightarrow[(r)]{p} \boxed{X} \xleftarrow[(s)]{q} \right\rangle\!\!\right\rangle = s \cdot D_{\frac{r}{r+s}}(p \,\|\, q)$$

*and* $$D_\alpha(p \,\|\, q) = \left\langle\!\!\left\langle \xrightarrow[(\frac{\alpha}{1-\alpha})]{p} \boxed{X} \xleftarrow{q} \right\rangle\!\!\right\rangle$$

However, the two classes are not identical, because the PDG divergences have extra limit points. One big difference is that the reverse KL divergence $D(q \,\|\, p)$ is not a Rényi divergence $D_\alpha(p \,\|\, q)$ for any value (or limit) of $\alpha$. This lack of symmetry has led others (e.g., Cichocki and Amari 2010) to work instead with a symmetric variant called $\alpha$-divergence, rescaled by an additional factor of $\frac{1}{\alpha}$. The relationships between these quantities can be seen in Figure 1.

The Chernoff divergence measures the tightest possible exponential bound on probability of error (Nielsen 2011) in Bayesian hypothesis testing. It also happens to be the smallest possible inconsistency of simultaneously believing $p$ and $q$, with total confidence 1.

**Corollary 10.2.** *The Chernoff Divergence between $p$ and $q$ equals*

$$\inf_{\beta \in (0,1)} \left\langle\!\!\left\langle \xrightarrow[(\beta)]{p} \boxed{X} \xleftarrow[(1-\beta)]{q} \right\rangle\!\!\right\rangle.$$

One significant consequence of representing divergences as inconsistencies is that we can use Lemma 1 to derive relationships between them. The following facts follow directly from Figure 1, by inspection.

**Corollary 10.3.**
1. *Rényi entropy is monotonic in its parameter $\alpha$.*
2. $D(p \,\|\, q) \geq 2D_B(p, q) \leq D(q \,\|\, p)$.
3. *If $q(p > 0) < 1$ (i.e., $q \not\ll p$), then $D(q \,\|\, p) = \infty$.*

These divergences correspond to PDGs with only two edges and one variable. What about more complex graphs? For a start, conditional divergences

$$D^{\mathrm{PDG}}_{(r,s)}\Big(p(Y|X) \,\Big\|\, q(Y|X) \,\Big|\, r(X)\Big) := \mathop{\mathbb{E}}_{x \sim r} D^{\mathrm{PDG}}_{(r,s)}\Big(p(Y|x) \,\Big\|\, q(Y|x)\Big)$$

can be represented straightforwardly as

$$D^{\mathrm{PDG}}_{(r,s)}(p \,\|\, q \,|\, r) = \left\langle\!\!\left\langle \xrightarrow[(\infty)]{r} \boxed{X} \overset{p\ (r)}{\underset{q\ (s)}{\rightrightarrows}} \boxed{Y} \right\rangle\!\!\right\rangle.$$

Other structures are useful intermediates. Lemma 1, plus some structural manipulation, gives visual proofs of many divergence properties; Figure 2 features such a proof of the data-processing inequality. And in general, PDG inconsistency can be viewed as a vast generalization of divergences to arbitrary structured objects.

# 6 VARIATIONAL OBJECTIVES AND BOUNDS

The fact that the incompatibility of $\mathcal{m}$ with a *specific* joint distribution $\mu$ is an upper bound on the inconsistency is not a deep one, but it is of a variational flavor. Here, we focus on the more surprising converse: PDG semantics capture general aspects of variational inference and provide a graphical proof language for it.

## 6.1 PDGs and Variational Approximations

We begin by recounting the standard development of the 'Evidence Lower BOund' (ELBO), a standard objective for training latent variable models (Blei, Kucukelbir, and McAuliffe 2017, §2.2). Suppose we have a model $p(X, Z)$, but only have access to observations of $x$. In service of adjusting $p(X, Z)$ to make our observations more likely, we would like to maximize $\log p(X{=}x)$, the "evidence" of $x$ (Proposition 4). Unfortunately, computing $p(X) = \sum_z p(X, Z{=}z)$ requires summing over all of $Z$, which can be intractable. The variational approach is as follows: fix a family of distributions $\mathcal{Q}$ that is easy to sample from, choose some $q(Z) \in \mathcal{Q}$, and define $\mathrm{ELBO}_{p,q}(x) := \mathbb{E}_{z\sim q} \log \frac{p(x,z)}{q(z)}$. This is something we can estimate, since we can sample from $q$. By Jensen's inequality,

$$\operatorname*{ELBO}_{p,q}(x) = \mathop{\mathbb{E}}_{q} \log \frac{p(x,Z)}{q(Z)} \leq \log \left[\mathop{\mathbb{E}}_{q} \frac{p(x,Z)}{q(Z)}\right] = \log p(x),$$

with equality if $q(Z) = p(Z)$. So to find $p$ maximizing $p(x)$, it suffices to adjust $p$ and $q$ to maximize $\mathrm{ELBO}_{p,q}(x)$,[4] provided $\mathcal{Q}$ is expressive enough.

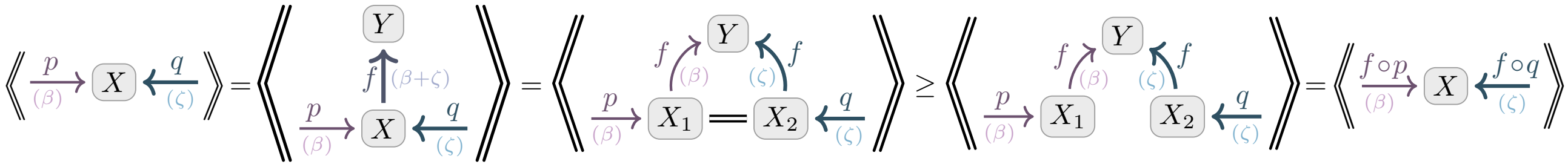

Figure 2: A visual proof of the data-processing inequality: $\boldsymbol{D}^{\mathrm{PDG}}_{(\beta,\zeta)}\big(p \,\big\|\, q\big) \geq \boldsymbol{D}^{\mathrm{PDG}}_{(\beta,\zeta)}\big(f \circ p \,\big\|\, f \circ q\big)$. In words: the cpd $f(Y|X)$ can always be satisfied, so adds no inconsistency. It is then equivalent to split $f$ and the variable $X$ into $X_1$ and $X_2$ with edges enforcing $X_1 = X_2$. But removing such edges can only decrease inconsistency. Finally, compose the remaining cpds to give the result. See the appendix for a full justification.

The formula for the ELBO is somewhat difficult to make sense of.[5] Nevertheless, it arises naturally as the inconsistency of the appropriate PDG.

link to proof

**Proposition 11.** *The negative ELBO of $x$ is the inconsistency of the PDG containing $p$,$q$, and $X = x$, with high confidence in $q$. That is,*

$-\mathrm{ELBO}_{p,q}(x) =$ 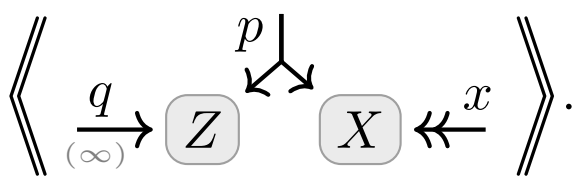

Owing to its structure, a PDG is often more intuitive and easier to work with than the formula for its inconsistency. To illustrate, we now give a simple and visually intuitive proof of the bound traditionally used to motivate the ELBO, via Lemma 1:

$\log \frac{1}{p(x)} =$ 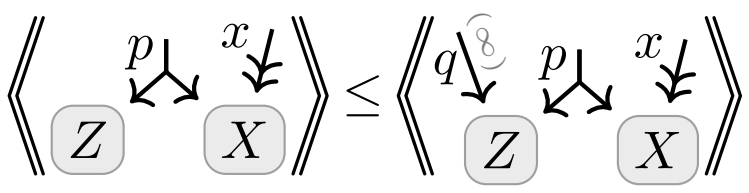 $= -\operatorname*{ELBO}_{p,q}(x).$

The first and last equalities are Propositions 4 and 11 respectively. Now to reap some pedagogical benefits. The second PDG has more edges so it is clearly at least as inconsistent. Furthermore, it's easy to see that equality holds when $q(Z) = p(Z)$: the best distribution for the left PDG has marginal $p(Z)$ anyway, so insisting on it incurs no further cost.

[4] or for many iid samples: $\max_{p,q} \sum_{x\in\mathcal{D}} \mathrm{ELBO}_{p,q}(x)$.

[5] Especially if $p, q$ are densities. See Appendix A.

### 6.2 Variational Auto-Encoders and PDGs

An autoencoder is a probabilistic model intended to compress a variable $X$ (e.g., an image) to a compact latent representation $Z$. Its structure is given by two conditional distributions: an encoder $e(Z|X)$, and a decoder $d(X|Z)$. Of course, not all pairs of cpds fill this role equally well. One important consideration is the *reconstruction error* (6): when we decode an encoded image, we would like it to resemble the original.

$$\mathrm{Rec}(x) := \mathop{\mathbb{E}}_{z\sim e(Z|x)} \underbrace{\mathrm{I}_{d(X|z)}(x)}_{\left(\substack{\text{additional bits required to}\\ \text{decode } x \text{ from its encoding } z}\right)} = \sum_z e(z\,|\,x) \log \frac{1}{d(x\,|\,z)} \tag{6}$$

There are other desiderata as well. Perhaps good latent representations $Z$ have uncorrelated components, and are normally distributed. We encode such wishful thinking as a belief $p(Z)$, known as a variational prior.

The data of a Variational Auto-Encoder (Kingma and Welling 2014; Rezende, Mohamed, and Wierstra 2014), or VAE, consists of $e(Z|X)$, $d(X|Z)$, and $p(Z)$. The encoder $e(Z|X)$ can be used as a variational approximation of $Z$, differing from $q(Z)$ of Section 6.1 only in that it can depend on $X$. VAEs are trained with the analogous form of the ELBO:

$$\begin{aligned}\mathrm{ELBO}_{p,e,d}(x) &:= \mathop{\mathbb{E}}_{z\sim e(Z|x)} \left[\log \frac{p(z)d(x\mid z)}{e(z\mid x)}\right] \\ &= -\,\mathrm{Rec}(x) - \boldsymbol{D}(e(Z|x)\parallel p).\end{aligned}$$

This gives us the following analog of Proposition 11.

**Proposition 12.** *The VAE loss of a sample $x$ is the inconsistency of the PDG comprising the encoder $e$ (with high confidence, as it defines the encoding), decoder $d$, prior $p$, and $x$. That is,*

link to proof

$-\mathrm{ELBO}_{p,e,d}(x) =$ 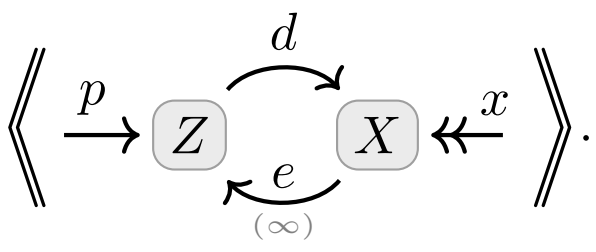

We now give a visual proof of the analogous variational bound. Let $\Pr_{p,d}(X,Z) := p(Z)d(X|Z)$ be the distri-bution that arises from decoding the prior. Then:

$$\log \frac{1}{\Pr_{p,d}(x)} =$$
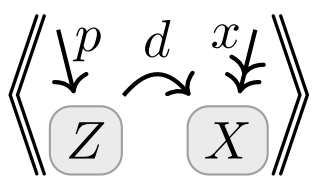

$$\leq$$
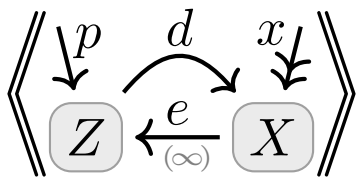

$$= \text{-ELBO}_{p,e,d}(x).$$

The first and last equalities are Propositions 4 and 12, and the inequality is Lemma 1. See the appendix for multi-sample analogs of the bound and Proposition 12.

### 6.3 The $\beta$-VAE Objective

The ELBO is not the only objective that has been used to train networks with a VAE structure. In the most common variant, due to Higgins et al. (2016), one weights the reconstruction error (6) and the 'KL term' differently, resulting in a loss function of the form

$$\beta\text{-ELBO}_{p,e,d}(x) := -\text{Rec}(x) - \beta \boldsymbol{D}(e(Z|x) \parallel p),$$

which, when $\beta = 1$, is the ELBO as before. The authors view $\beta$ as a regularization strength, and argue that it sometimes helps to have a stronger prior. Sure enough:

link to proof

**Proposition 13.** *$-\beta$-ELBO$_{p,e,d}(x)$ is the inconsistency of the same PDG, but with confidence $\beta$ in $p(Z)$.*

## 7 FREE ENERGY AND INCONSISTENCY

A weighted factor graph $\Psi = (\phi_J, \theta_J)_{J \in \mathcal{J}}$, where each $\theta_J$ is a real-valued weight, $J$ is associated with a subset of variables $\mathbf{X}_J$, and $\phi_J : \mathcal{V}(\mathbf{X}_J) \to \mathbb{R}$, determines a distribution by

$$\Pr_\Psi(\mathbf{x}) = \frac{1}{Z_\Psi} \prod_{J \in \mathcal{J}} \phi_J(\mathbf{x}_J)^{\theta_J}.$$

$Z_\Psi$ is the constant $\sum_{\mathbf{x}} \prod_{J \in \mathcal{J}} \phi_J(\mathbf{x}_J)^{\theta_J}$ required to normalize the distribution, and is known as the *partition function*. Computing $\log Z_\Psi$ is intimately related to probabilistic inference in factor graphs (Ma et al. 2013). Following Richardson and Halpern (2021), let $\mathcal{M}_\Psi$ be the PDG with edges $\{ \xrightarrow{J} \mathbf{X}_J \}_{\mathcal{J}}$, cpds $p_J(\mathbf{X}_J) \propto \phi_J(\mathbf{X}_J)$, and weights $\alpha_J, \beta_J := \theta_J$. There, it is shown that $\Pr_\Psi$ is the unique minimizer of $[\![\mathcal{M}_\Psi]\!]_1$. But what about the corresponding inconsistency, $\langle\!\langle \mathcal{M}_\Psi \rangle\!\rangle_1$?

If the factors are normalized and all variables are edge targets, then $Z_\Psi \leq 1$, so $\log \frac{1}{Z_\Psi} \geq 0$ measures how far the product of factors is from being a probability distribution. So in a sense, it measures $\Psi$'s inconsistency.

link to proof

**Proposition 14.** *For all weighted factor graphs $\Psi$, we have that $\langle\!\langle \mathcal{M}_\Psi \rangle\!\rangle_1 = -\log Z_\Psi$.*

The exponential families generated by weighted factor graphs are a cornerstone of statistical mechanics, where $-\log Z_\Psi$ is known as the (Heimholz) free energy. It is also an especially natural quantity to minimize: the principle of free-energy minimization has been enormously succesful in describing of not only chemical and biological systems (Chipot and Pohorille 2007), but also cognitive ones (Friston 2009).

## 8 BEYOND STANDARD LOSSES: A CONCRETE EXAMPLE

In contexts where a loss function is standard, it is usually for good reason—which is why we have focused on recovering standard losses. But most situations are non-standard, and even if they have standard subcomponents, those components may interact with one another in more than one way. Correspondingly, there is generally more than one way to cobble standard loss functions together. How should you choose between them? By giving a principled model of the situation.

Suppose we want to train a predictor network $h(Y|X)$ from two sources of information: partially corrupted data with distribution $d(X,Y)$, and a simulation with distribution $s(X,Y)$. If the simulation is excellent and the data unsalvagable, we would have high confidence in $s$ and low confidence in $d$, in which case we would train with cross entropy with respect to $s$, $\mathcal{L}_{\texttt{sim}} := \mathbb{E}_s[\log {}^1\!/_{h(Y|X)}]$. Conversely, if the simulation were bad and the data mostly intact, we would use $\mathcal{L}_{\texttt{dat}}$, the cross entropy with respect to $d$. What if we're not so confident in either?

One approach a practitioner might find attractive is to make a dataset from samples of both $s$ and $d$, or equivalently, train with a convex combination of the two previous losses, $\mathcal{L}_1 := \lambda_{\texttt{s}} \mathcal{L}_{\texttt{sim}} + \lambda_{\texttt{d}} \mathcal{L}_{\texttt{dat}}$ for some $\lambda_{\texttt{s}}, \lambda_{\texttt{d}} > 0$ with $\lambda_{\texttt{s}} + \lambda_{\texttt{d}} = 1$. This amounts to training $h$ with cross entropy with respect to the mixture $\lambda_{\texttt{s}} s + \lambda_{\texttt{d}} d$. Doing so treats $d$ and $s$ as completely unrelated, and so redundancy is not used to correct errors—a fact on display when we present the modeling choices in PDG form, such as

$$\mathcal{L}_1 =$$
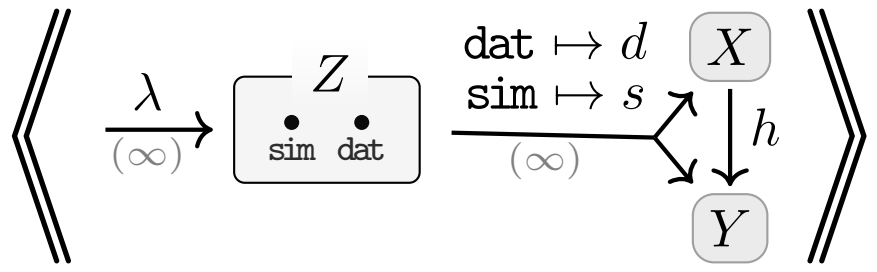

in which a swich variable $Z$ with possible values $\{\texttt{sim}, \texttt{dat}\}$ controls whether samples come from $s$ or $d$, and is distributed according to $\lambda(Z = \texttt{sim}) = \lambda_{\texttt{s}}$.

Our practitioner now tries a different approach: draw data samples $(x, y) \sim d$ but discount $h$'s surprisal when the simulator finds the point unlikely, via loss $\mathcal{L}_2 := \mathbb{E}_d[s(X,Y) \log {}^1\!/_{h(Y|X)}]$. This is the cross entropy with respect to the (unnormalized) product density $ds$, which in many ways is appropriate. However, by this

metric, the optimal predictor $h^*(Y|x) \propto d(Y|x)s(Y|x)$ is *uncalibrated* (Dawid 1982). If the data and simulator agree ($d{=}s$), then we would want $h(Y|x){=}s(Y|x)$ for all $x$, but instead we get $h^*(Y|x) \propto s(Y|x)^2$. So $h^*$ is overconfident. What went wrong? $\mathcal{L}_2$ cannot be written as the (ordinary $\gamma{=}0$) inconsistency of a PDG containing only $s, h$, and $d$, but for a large fixed $\gamma$, it is essentially the $\gamma$-inconsistency

$\mathcal{L}_2 \approx C$ 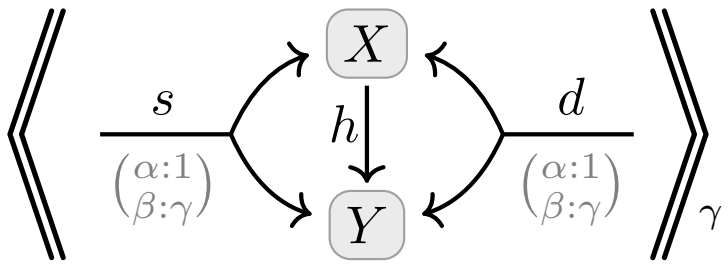 $+\ const,$

where $C$ is the constant required to normalize the joint density $sd$, and *const* does not depend on $h$. However, the values of $\boldsymbol{\alpha}$ in this PDG indicate an over-determination of $XY$ (it is determined in two different ways), and so $h^*$ is more deterministic than intended. By contrast,

$\mathcal{L}_3 :=$ 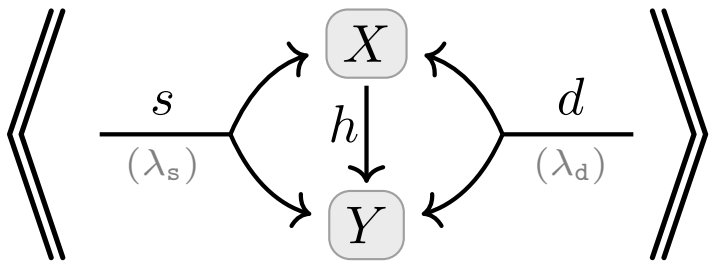 ,

does not have this issue: the optimal predictor $h^*$ according to $\mathcal{L}_3$ is proportional to the $\lambda$-weighted geometric mean of $s$ and $d$. It seems that our approach, in addition to providing a unified view of standard loss functions, can also suggest more appropriate loss functions in practical situations.

## 9 REVERSE-ENGINEERING LOSS?

Given an *arbitrary* loss function, can we find a PDG that gives rise to it? The answer appears to be yes—although not without making unsavory modeling choices. Without affecting its semantics, one may add the variable $\mathtt{T}$ that takes values $\{\mathtt{t}, \mathtt{f}\}$, and the event $\mathtt{T} = \mathtt{t}$, to any PDG. Now, given a cost function $c : \mathcal{V}(X) \to \mathbb{R}_{\geq 0}$, define the cpd $\hat{c}(\mathtt{T}|X)$ by $\hat{c}(\mathtt{t}|x) := e^{-c(x)}$. By threatening to generate the falsehood $\mathtt{f}$ with probability dependent on the cost of $X$, $\hat{c}$ ties the value of $X$ to inconsistency.

link to proof

**Proposition 15.** $\left\langle\!\!\left\langle \xrightarrow[(\infty)]{p} X \xrightarrow{\hat{c}} \mathtt{T} \twoheadleftarrow{\mathtt{t}} \right\rangle\!\!\right\rangle = \mathop{\mathbb{E}}_{x\sim p} c(x).$

Setting confidence $\beta_p := \infty$ may not be realistic since we're still training the model $p$, but doing so is necessary to recover $\mathbb{E}_p\, c$.[6] Any mechanism that generates inconsistency based on the value of $X$ (such as this one) also works in reverse: the PDG "squirms", contorting the probability of $X$ to disperse the inconsistency. One cannot cannot simply "emit loss" without affecting the rest of the model, as one does with utility in an Influence Diagram (Howard 1983). Even setting every $\beta := \infty$ may not be enough to prevent the squirming. To illustrate, consider a model $\mathcal{S}$ of the supervised learning setting (predict $Y$ from $X$), with labeled data $\mathcal{D}$, model $h$, and a loss function $\ell$ on pairs of output labels.

Concretely, define:

$\mathcal{S} :=$ 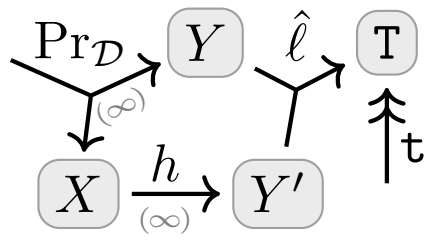 and $\mathcal{L} := \mathop{\mathbb{E}}_{\substack{(x,y)\sim \Pr_\mathcal{D} \\ y' \sim p(Y'\,|\,x)}} \big[\ell(y, y')\big].$

Given Proposition 15, one might imagine $\langle\!\langle \mathcal{S} \rangle\!\rangle = \mathcal{L}$, but this is not so. In some ways, $\langle\!\langle \mathcal{S} \rangle\!\rangle$ is actually preferable. The optimal $h(Y'|X)$ according to $\mathcal{L}$ is a degenerate cpd that places all mass on the label(s) $y^*_X$ minimizing expected loss, while the optimal $h(Y'|X)$ according to $\langle\!\langle \mathcal{S} \rangle\!\rangle$ is $\Pr_\mathcal{D}(Y|X)$, which means that it is calibrated, unlike $\ell$. If, in addition, we set $\alpha_p, \alpha_{\Pr_\mathcal{D}} := 1$ and strictly enforce the qualitative picture, finally no more squirming is possible, as we arrive at $\lim_{\gamma\to\infty} \langle\!\langle \mathcal{S} \rangle\!\rangle_\gamma = \mathcal{L}$.

In the process, we have given up our ability to tolerate inconsistency by setting all probabilistic modeling choices in stone. What's more, we've dragged in the global parameter $\gamma$, further handicapping our ability to compose this model with others. To summarize: while model inconsistency readily generates appropriate loss functions, the converse does not work as well. Reverse-enerineering a loss may require making questionable modeling choices with absolute certainty, resulting in brittle models with limited potential for composition. In the end, we must confront our modeling choices; good loss functions come from good models.

## 10 FINAL REMARKS

We seen that that PDG semantics, in the same stroke by which they capture Bayesian Networks and Factor Graphs (Richardson and Halpern 2021), also generate many standard loss functions, including some non-trivial ones. In each case, the appropriate loss arises simply by articulating modeling assumptions, and then measuring inconsistency. Viewing loss functions in this way also has beneficial side effects, including an intuitive visual proof language for reasoning about the relationships between them.

This "universal loss", which provides a principled way of choosing an optimization objective, may be of particular interest to the AI alignment community.

[6] If $\beta_p$ were instead equal to 1, we would have obtained $-\log \mathbb{E}_p \exp(-c(X))$, with optimal distribution $\mu(X) \neq p(X)$.

## Acknowledgements

Work supported in part by MURI grant W911NF-19-1-0217. Many thanks to my advisor, Joe Halpern, for his generous support, and for valuable critiques of many drafts. Thanks as well to my reviewers, who pushed me to better explain the confidence parameters, and to include a practical example (Section 8). Finally, thanks to my friends, particularly Varsha Kishore and Greg Yauney, for helping me to refine the presentation of these ideas.

# A THE FINE PRINT FOR PROBABILITY DENSITIES

**Densities and Masses.** Many of our results (Propositions 2 to 5, 11, 12, 17, 19 and 20) technically require the distribution to be represented with a mass function (as opposed to a probability density function, or pdf). A PDG containg both pdf and a finitely supported distribution on the same variable will typically have infinite inconsistency—but this is not just a quirk of the PDG formalism.

Probability density is not dimensionless (like probability mass), but rather has inverse $X$-units (e.g., probability per meter), so depends on an arbitrary choice of scale (the pdf for probability per meter and per centimeter will yield different numbers). In places where the objective does not have units that cancel before we take a logarithm, the use of a probability density $p(X)$ becomes sensitive to this arbitrary choice of parameterization. For instance, the analog of surprisal, $-\log p(x)$ for a pdf $p$, or its expectation, called differential entropy, both depend on an underlying scheme of measurement (an implicit base measure).

On the other hand, this choice of scale ultimately amounts to an additive constant. Moreover, beyond a certain point, decreasing the discretization size $k$ of a discretized approximation $\tilde{p}_k(X)$ *also* contributes a constant that depends only on $k$. But such constants are irrelevant for optimization, and so, even though such quantities are ill-defined and arguably meaningless in the continuous limit, the use of the continuous analogs as loss functions is still justified.

The bottom line is that all our results hold in a uniform way for every discretization size — yet in the limit as the discretization becomes smaller, an inconsistency may diverge to infinity. However, this divergence stems from an additive constant that depends only on the discretization size, which is irrelevant to its employment as a loss function. As a result, using one of these "unbalanced" functions involving densities where the units do not work out properly, results in a morally equivalent loss function, except without a diverging constant.

**Markov Kernels.** In the more general setting of measurable spaces, one may want to adjust the definition of a cpd that we gave, so that one instead works with *Markov Kernels.* This imposes an additional constraint: suppose the variable $Y$ takes values in the measurable space $(\mathcal{V}(Y), \mathcal{B})$. If $p(Y|X)$ is to be a *Markov Kernel*, then for every fixed measurable subset $B \in \mathcal{B}$ of the measure space, the we must require that $x \mapsto \Pr(B|x)$ be a measurable function (with respect to the measure space in which $X$ takes values). This too mostly does not bear on the present discussion, because the $\sigma$-algebras for all measure spaces of interest, are fine enough that one can get an arbitrarily close approximation of any cpd with a Markov Kernels. This means that the infemum defining the inconsistency of a PDG does not change.

# B FURTHER RESULTS AND GENERALIZATIONS

## B.1 Full Characterization of Gaussian Predictors

The inconsistency of a PDG containing two univariate Gaussian regressors of with arbitrary paremeters and confidences, is most cleanly articulated in terms of the geometric and quadratic means.

**Definition 2** (Weighted Power Mean)**.** The weighted power mean $\mathrm{M}_p^w(\mathbf{r})$ of the collection of real numbers $\mathbf{r} = r_1, \ldots, r_n$ with respect to the convex weights $w = w_1, \ldots, w_n$ satisfying $\sum_i w_i = 1$, is given by

$$\mathrm{M}_p^w(\mathbf{r}) := \Big( \sum_{i=1}^n w_i (r_i)^p \Big)^{\frac{1}{p}}.$$

We omit the superscript as a shorthand for the uniform weighting $w_i = {}^1\!/_N$. □

Many standard means, such as those in Table 1, are special cases. It is well known that $\mathrm{M}_p^w(\mathbf{r})$ is increasing in $p$, and strictly so if not all elements of $\mathbf{r}$ are identical. In particular, $\mathrm{QM}_w(a,b) > \mathrm{GM}_w(a,b)$ for all $a \neq b$ and positive weights $w$. We now present the result.

| Name | $p$ | Formula |
|---|---|---|
| Harmonic | $(p=-1)$: | $\mathrm{HM}_w(\mathbf{r}) = 1 / \left(\sum_{i=1}^n w_i / r_i\right)$ |
| Geometric | $(\lim p \to 0)$: | $\mathrm{GM}_w(\mathbf{r}) = \prod_{i=1}^n r_i^{w_i}$ |
| Arithmetic | $(p=1)$: | $\mathrm{AM}_w(\mathbf{r}) = \sum_{i=1}^n w_i r_i$ |
| Quadratic | $(p=2)$: | $\mathrm{QM}_w(\mathbf{r}) = \sqrt{\sum_{i=1}^n w_i r_i^2}$ |

Table 1: special cases of the $p$-power mean $\mathrm{M}_p^w(\mathbf{r})$

link to proof

**Proposition 16.** *Consider a PDG containing two (distinct) conditional Gaussian distributions on a variable $Y$, whose parameters can both depend on a variable $X$. Its inconsistency takes the form*

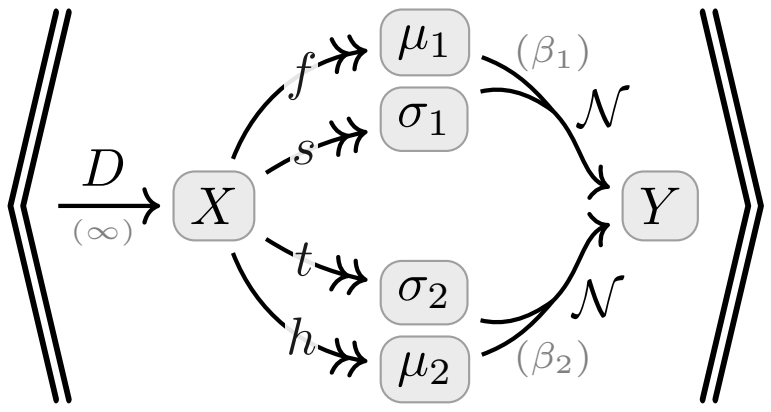

$$\langle\!\langle \cdots \rangle\!\rangle = \mathop{\mathbb{E}}_{D}\left[ (\beta_1+\beta_2) \log \frac{\mathrm{QM}_{\hat\beta}(\sigma_1,\sigma_2)}{\mathrm{GM}_{\hat\beta}(\sigma_1,\sigma_2)} + \frac{1}{2}\frac{\beta_1\beta_2}{\beta_1+\beta_2}\left(\frac{\mu_1-\mu_2}{\mathrm{QM}_{\hat\beta}(\sigma_1,\sigma_2)}\right)^2\right] \tag{7}$$

$$= \frac{1}{2}\mathop{\mathbb{E}}_{x\sim D}\left[\frac{\beta_1\beta_2}{2}\frac{\big(f(x)-h(x)\big)^2}{\beta_2 s(x)^2+\beta_1 t(x)^2} + \frac{\beta_1+\beta_2}{2}\log\frac{\beta_2 s(x)^2+\beta_1 t(x)^2}{\beta_1+\beta_2} \begin{array}{l} -\beta_2 \log s(x) \\ -\beta_1 \log t(x) \end{array}\right]$$

*where $\hat\beta = (\frac{\beta_2}{\beta_1+\beta_2}, \frac{\beta_1}{\beta_1+\beta_2})$ represents the normalized and reversed vector of conficences $\beta = (\beta_1, \beta_2)$ for the two distributions, and $\mu_1 = f(X)$, $\mu_2 = g(X)$, $\sigma_1 = s(X)$, $\sigma_2 = t(X)$ are random variables over $X$.*

The PDG on the left is semantically equivalent to (and in particular has the same inconsistency as) the PDG

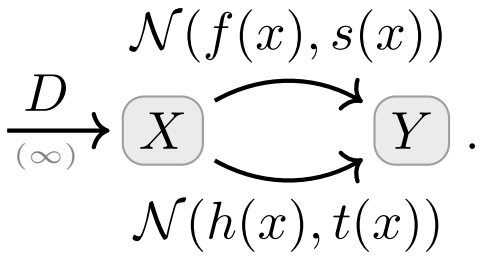

This illustrates an orthogonal point: that PDGs handle composition of functions as one would expect, so that it is equivalent to model an entire process as a single arrow, or to break it into stages, ascribing an arrow to each stage, with one step of randomization.

As a bonus, Proposition 16 also gives a proof of the inequality of the weighted geometric and quadratic means.

**Corollary 16.1.** *For all $\sigma_1$ and $\sigma_2$, and all weight vectors $\beta$, $\mathrm{QM}_{\hat\beta}(\sigma_1,\sigma_2) \geq \mathrm{GM}_{\hat\beta}(\sigma_1,\sigma_2)$.*

## B.2 Full-Dataset ELBO and Bounds

We now present the promised multi-sample analogs from Section 6.1.

link to proof

**Proposition 17.** *The following analog of Proposition 12 for a whole dataset $\mathcal{D}$ holds:*

$$-\mathop{\mathbb{E}}_{\Pr_{\mathcal{D}}} \mathrm{ELBO}_{p,e,d}(X) = \langle\!\langle \cdots \rangle\!\rangle + \mathrm{H}(\Pr_{\mathcal{D}}).$$

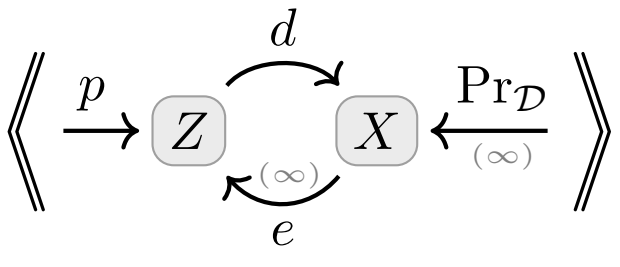

Propositions 3 and 17 then give us an analog of the visual bounds in the body of the main paper (Section 6.1)

for many i.i.d. datapoints at once, with only a single application of the inequality:

$$-\log \Pr(\mathcal{D}) = -\log \prod_{i=1}^{m} \Big(\Pr(x^{(i)})\Big) = -\frac{1}{m}\sum_{i=1}^{m} \log \Pr(x^{(i)}) =$$

$$\mathrm{H}(\Pr_{\mathcal{D}}) + \left\langle\!\!\left\langle \right.\right.$$ 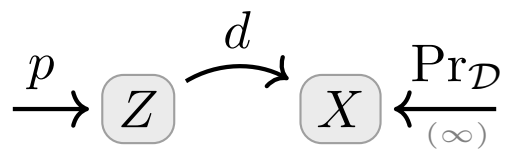 $$\left.\left.\right\rangle\!\!\right\rangle \leq \left\langle\!\!\left\langle \right.\right.$$ 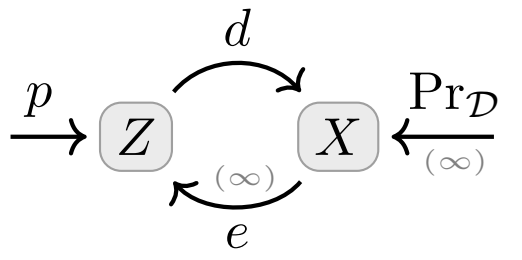 $$\left.\left.\right\rangle\!\!\right\rangle + \mathrm{H}(\Pr_{\mathcal{D}})$$

$$= -\mathop{\mathbb{E}}_{\Pr_{\mathcal{D}}} \mathop{\mathrm{ELBO}}_{p,e,d}(X)$$

We also have the following formal statement of Proposition 13.

**Proposition 18.** *The negative $\beta$-ELBO objective for a prior $p(X)$, encoder $e(Z \mid X)$, decoder $d(X \mid Z)$, at a sample $x$, is equal to the inconsistency of the corresponding PDG, where the prior has confidence equal to $\beta$. That is,* [link to proof]

$$-\beta\text{-ELBO}_{p,e,d}(x) = \left\langle\!\!\left\langle \right.\right.$$ 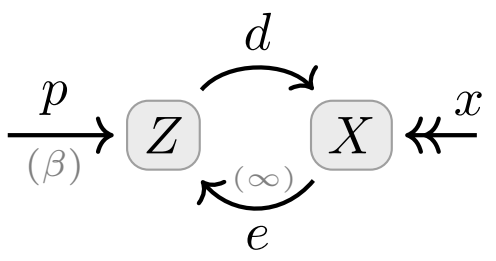 $$\left.\left.\right\rangle\!\!\right\rangle$$

As a specific case (i.e., effectively by setting $\beta_p := 0$), we get the reconstruction error as the inconsistency of an autoencoder (without a variational prior):

**Corollary 18.1** (reconstruction error as inconsistency)**.**

$$-\mathrm{Rec}_{ed,d}(x) := \mathop{\mathbb{E}}_{z \sim e(Z|x)} \mathrm{I}_{d(X|z)}(x) = \left\langle\!\!\left\langle \right.\right.$$ 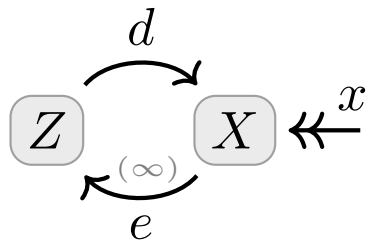 $$\left.\left.\right\rangle\!\!\right\rangle$$

### B.3 More Variants of Cross Entropy Results

First, we show that our cross entropy results hold for all $\gamma$, in the sense that $\gamma$ contributes only a constant.

**Proposition 19.** *Given a model determining a probability distribution with mass function $p(X)$, and samples $\mathcal{D} = \{x_i\}_{i=1}^m$ determining an empirical distribution $\Pr_{\mathcal{D}}$, the following are equal, for all $\gamma \geq 0$:* [link to proof]

1. *The average negative log likelihood* $\ell(p; \mathcal{D}) = -\frac{1}{m}\sum_{i=1}^{m} \log p(x_i)$
2. *The cross entropy of $p$ relative to* $\Pr_{\mathcal{D}}$
3. $[\![ p ]\!]_\gamma(\Pr_{\mathcal{D}}) \quad +(1+\gamma)\,\mathrm{H}(\Pr_{\mathcal{D}})$
4. $\left\langle\!\!\left\langle \right.\right.$ 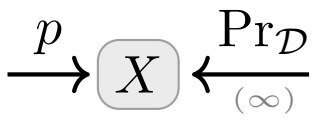 $\left.\left.\right\rangle\!\!\right\rangle_\gamma \quad +(1+\gamma)\,\mathrm{H}(\Pr_{\mathcal{D}})$

As promised, we now give the simultaneous generalization of the surprisal result (Proposition 2) to both multiple samples (like in Proposition 3) and partial observations (as in Proposition 4).

**Proposition 20.** *The average* marginal *negative log likelihood $\ell(p; x) := -\frac{1}{|\mathcal{D}|}\sum_{x \in \mathcal{D}} \log \sum_z p(x, z)$ is the inconsistency of the PDG containing $p$ and the data distribution $\Pr_{\mathcal{D}}$, plus the entropy of the data distribution (which is constant in $p$). That is,* [link to proof]

$$\ell(p; \mathcal{D}) = \left\langle\!\!\left\langle \right.\right.$$ 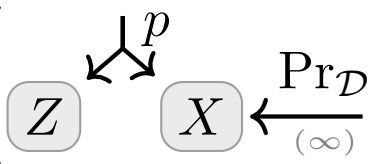 $$\left.\left.\right\rangle\!\!\right\rangle + \mathrm{H}(\Pr_{\mathcal{D}}).$$

# C PROOFS

**Lemma 1.** *Suppose PDGs $\mathcal{M}$ and $\mathcal{M}'$ differ only in their edges (resp. $\mathcal{E}$ and $\mathcal{E}'$) and confidences (resp. $\boldsymbol{\beta}$ and $\boldsymbol{\beta}'$). If $\mathcal{E} \subseteq \mathcal{E}'$ and $\beta_L \leq \beta'_L$ for all $L \in \mathcal{E}$, then $\langle\!\langle \mathcal{M} \rangle\!\rangle_\gamma \leq \langle\!\langle \mathcal{M}' \rangle\!\rangle_\gamma$ for all $\gamma$.*

*Proof.* For every $\mu$, adding more edges only adds non-negative terms to (1), while increasing $\beta$ results in larger coefficients on the existing (non-negative) terms of (1). So for every fixed distribution $\mu$, we have $[\![\mathcal{M}]\!]_\gamma(\mu) \leq [\![\mathcal{M}']\!]_\gamma(\mu)$. So it must also be the case that the infemum over $\mu$, so we find that $\langle\!\langle \mathcal{M} \rangle\!\rangle \leq \langle\!\langle \mathcal{M}' \rangle\!\rangle$. □

**Proposition 2.** *Consider a distribution $p(X)$. The inconsistency of the PDG comprising $p$ and $X{=}x$ equals the surprisal $\mathrm{I}_p[X{=}x]$. That is,*

$$\mathrm{I}_p[X{=}x] = \left\langle\!\!\left\langle \xrightarrow{p} \boxed{X} \overset{x}{\twoheadleftarrow} \right\rangle\!\!\right\rangle.$$

*(Recall that $\langle\!\langle \mathcal{M} \rangle\!\rangle$ is the inconsistency of the PDG $\mathcal{M}$.)*

*Proof.* Any distribution $\mu(X)$ that places mass on some $x' \neq x$ will have infinite KL divergence from the point mass on $x$. Thus, the only possibility for a finite consistency arises when $\mu = \delta_x$, and so

$$\left\langle\!\!\left\langle \xrightarrow{p} \boxed{X} \overset{x}{\twoheadleftarrow} \right\rangle\!\!\right\rangle = \left[\!\!\left[ \xrightarrow{p} \boxed{X} \overset{x}{\twoheadleftarrow} \right]\!\!\right](\delta_x) = \boldsymbol{D}(\delta_x \parallel p) = \log \frac{1}{p(x)} = \mathrm{I}_p(x).$$

□

Proposition 19 is a generalization of Proposition 3, so we prove them at the same time.

**Proposition 3.** *If $p(X)$ is a probabilistic model of $X$, and $\mathcal{D} = \{x_i\}_{i=1}^m$ is a dataset with empirical distribution $\Pr_\mathcal{D}$, then* $\mathrm{CrossEntropy}(\Pr_\mathcal{D}, p) =$

$$\frac{1}{m}\sum_{i=1}^m \mathrm{I}_p[X{=}x_i] = \left\langle\!\!\left\langle \xrightarrow{p} \boxed{X} \xleftarrow[(\infty)]{\Pr_\mathcal{D}} \right\rangle\!\!\right\rangle + \mathrm{H}(\Pr_\mathcal{D}).$$

**Proposition 19.** *Given a model determining a probability distribution with mass function $p(X)$, and samples $\mathcal{D} = \{x_i\}_{i=1}^m$ determining an empirical distribution $\Pr_\mathcal{D}$, the following are equal, for all $\gamma \geq 0$:*

1. *The average negative log likelihood $\ell(p; \mathcal{D}) = -\frac{1}{m}\sum_{i=1}^m \log p(x_i)$*
2. *The cross entropy of $p$ relative to $\Pr_\mathcal{D}$*
3. $[\![\, p \,]\!]_\gamma(\Pr_\mathcal{D}) \quad + (1+\gamma)\,\mathrm{H}(\Pr_\mathcal{D})$
4. $\left\langle\!\!\left\langle \xrightarrow{p} \boxed{X} \xleftarrow[(\infty)]{\Pr_\mathcal{D}} \right\rangle\!\!\right\rangle_\gamma \quad + (1+\gamma)\,\mathrm{H}(\Pr_\mathcal{D})$

*Proof.* The equality of 1 and 2 is standard. The equality of 3 and 4 can be seen by the fact that in the limit of infinite confidence on $\Pr_\mathcal{D}$, the optimal distribution must also equal $\Pr_\mathcal{D}$, so the least inconsistency is attained at this value. Finally it remains to show that the first two and last two are equal:

$$\begin{aligned} [\![\, p \,]\!]_\gamma(\Pr_\mathcal{D}) + (1+\gamma)\,\mathrm{H}(\Pr_\mathcal{D}) &= \boldsymbol{D}(\Pr_\mathcal{D} \parallel p) - \gamma\,\mathrm{H}(\Pr_\mathcal{D}) + (1+\gamma)\,\mathrm{H}(\Pr_\mathcal{D}) \\ &= \boldsymbol{D}(\Pr_\mathcal{D} \parallel p) + \mathrm{H}(\Pr_\mathcal{D}) \end{aligned}$$

$$= \mathbb{E}_{\Pr_\mathcal{D}}\left[\log \frac{\Pr_\mathcal{D}}{p} + \log \frac{1}{\Pr_\mathcal{D}}\right] = \mathbb{E}_{\Pr_\mathcal{D}}\left[\log \frac{1}{p}\right],$$

which is the cross entropy, as desired. ☐

**Proposition 4.** *If $p(X,Z)$ is a joint distribution, then the information content of the partial observation $X=x$ is given by*

$$\mathrm{I}_p[X{=}x] = \left\langle\!\!\left\langle \boxed{Z} \overset{p}{\curlywedge} \boxed{X} \overset{x}{\twoheadleftarrow} \right\rangle\!\!\right\rangle. \tag{2}$$

*Proof.* As before, all mass of $\mu$ must be on $x$ for it to have a finite score. Thus it suffices to consider joint distributions of the form $\mu(X,Z) = \delta_x(X)\mu(Z)$. We have

$$\begin{aligned}
\left\langle\!\!\left\langle \boxed{Z} \overset{p}{\curlywedge} \boxed{X} \overset{x}{\twoheadleftarrow} \right\rangle\!\!\right\rangle &= \inf_{\mu(Z)} \left[\!\!\left[ \boxed{Z} \overset{p}{\curlywedge} \boxed{X} \overset{x}{\twoheadleftarrow} \right]\!\!\right] \Big(\delta_x(X)\mu(Z)\Big) \\
&= \inf_{\mu(Z)} \boldsymbol{D}\Big(\delta_x(X)\mu(Z) \;\Big\|\; p(X,Z)\Big) \\
&= \inf_{\mu(Z)} \mathop{\mathbb{E}}_{z\sim\mu} \log \frac{\mu(z)}{p(x,z)} \;=\; \inf_{\mu(Z)} \mathop{\mathbb{E}}_{z\sim\mu} \log \frac{\mu(z)}{p(x,z)} \frac{p(x)}{p(x)} \\
&= \inf_{\mu(Z)} \mathop{\mathbb{E}}_{z\sim\mu} \left[\log \frac{\mu(z)}{p(z\mid x)} + \log \frac{1}{p(x)}\right] \\
&= \inf_{\mu(Z)} \Big[\boldsymbol{D}(\mu(Z) \parallel p(Z\mid x))\Big] + \log \frac{1}{p(x)} \\
&= \log \frac{1}{p(x)} = \mathrm{I}_p(x) && \text{[Gibbs Inequality]}
\end{aligned}$$

☐

**Proposition 20.** *The average* marginal *negative log likelihood $\ell(p;x) := -\frac{1}{|\mathcal{D}|}\sum_{x\in\mathcal{D}} \log \sum_z p(x,z)$ is the inconsistency of the PDG containing $p$ and the data distribution $\Pr_\mathcal{D}$, plus the entropy of the data distribution (which is constant in $p$). That is,*

$$\ell(p;\mathcal{D}) = \left\langle\!\!\left\langle \boxed{Z} \overset{p}{\curlywedge} \boxed{X} \xleftarrow[(\infty)]{\Pr_\mathcal{D}} \right\rangle\!\!\right\rangle + \mathrm{H}(\Pr_\mathcal{D}).$$

*Proof.* The same idea as in Proposition 4, but a little more complicated.

$$\begin{aligned}
\left\langle\!\!\left\langle \boxed{Z} \overset{p}{\curlywedge} \boxed{X} \overset{\Pr_\mathcal{D}!}{\twoheadleftarrow} \right\rangle\!\!\right\rangle &= \inf_{\mu(Z\mid X)} \left[\!\!\left[ \boxed{Z} \overset{p}{\curlywedge} \boxed{X} \overset{\Pr_\mathcal{D}!}{\twoheadleftarrow} \right]\!\!\right] \Big(\Pr_\mathcal{D}(X)\mu(Z\mid X)\Big) \\
&= \inf_{\mu(Z\mid X)} \boldsymbol{D}\Big(\Pr_\mathcal{D}(X)\mu(Z\mid X) \;\Big\|\; p(X,Z)\Big) \\
&= \inf_{\mu(Z\mid X)} \mathop{\mathbb{E}}_{\substack{x\sim\Pr_\mathcal{D}\\ z\sim\mu}} \log \frac{\mu(z\mid x)\Pr_\mathcal{D}(x)}{p(x,z)} \\
&= \frac{1}{|\mathcal{D}|} \inf_{\mu(Z\mid X)} \sum_{x\in\mathcal{D}} \mathop{\mathbb{E}}_{z\sim\mu(Z\mid x)} \log \frac{\mu(z\mid x)\Pr_\mathcal{D}(x)}{p(x,z)} \frac{p(x)}{p(x)}
\end{aligned}$$

$$= \frac{1}{|\mathcal{D}|} \inf_{\mu(Z|X)} \sum_{x \in \mathcal{D}} \left[ \mathop{\mathbb{E}}_{z\sim\mu} \left[ \log \frac{\mu(z \mid x)}{p(z \mid x)} \right] + \log \frac{1}{p(x)} - \log \frac{1}{\Pr_{\mathcal{D}}(x)} \right]$$

$$= \frac{1}{|\mathcal{D}|} \sum_{x \in \mathcal{D}} \left[ \inf_{\mu(Z|x)} \mathop{\mathbb{E}}_{z\sim\mu} \left[ \log \frac{\mu(z \mid x)}{p(z \mid x)} \right] + \log \frac{1}{p(x)} - \log \frac{1}{\Pr_{\mathcal{D}}(x)} \right]$$

$$= \frac{1}{|\mathcal{D}|} \sum_{x \in \mathcal{D}} \left[ \inf_{\mu(Z)} \Big[ \boldsymbol{D}(\mu(Z) \parallel p(Z \mid x)) \Big] + \log \frac{1}{p(x)} \right] - \mathrm{H}(\Pr_{\mathcal{D}})$$

$$= \frac{1}{|\mathcal{D}|} \sum_{x \in \mathcal{D}} \log \frac{1}{p(x)} - \mathrm{H}(\Pr_{\mathcal{D}})$$

$$= \frac{1}{|\mathcal{D}|} \sum_{x \in \mathcal{D}} \mathrm{I}_p(x) - \mathrm{H}(\Pr_{\mathcal{D}})$$

$$\Big( \quad = \boldsymbol{D}(\Pr_{\mathcal{D}} \parallel p) \quad \Big)$$

□

**Proposition 5.** *The inconsistency of the PDG comprising a probabilistic predictor $h(Y|X)$, and a high-confidence empirical distribution $\Pr_{\mathcal{D}}$ of a dataset $\mathcal{D} = \{(x_i, y_i)\}_{i=1}^m$ equals the cross-entropy loss (minus the empirical uncertainty in $Y$ given $X$, a constant depending only on $\mathcal{D}$). That is,*

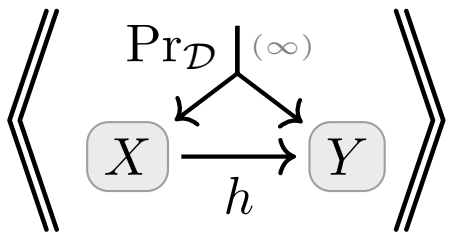

$$= \frac{1}{m} \sum_{i=1}^{m} \log \frac{1}{h(y_i \mid x_i)} - \mathrm{H}_{\Pr_{\mathcal{D}}}(Y|X).$$

*Proof.* $\Pr_{\mathcal{D}}$ has high confidence, it is the only joint distribution $\mu$ with finite score. Since $f$ is the only other edge, the inconsistency is therefore

$$\mathop{\mathbb{E}}_{x\sim\Pr_{\mathcal{D}}} \boldsymbol{D}\Big(\Pr_{\mathcal{D}}(Y \mid x) \,\Big\|\, f(Y \mid x)\Big) = \mathop{\mathbb{E}}_{x,y\sim\Pr_{\mathcal{D}}} \left[ \log \frac{\Pr_{\mathcal{D}}(y \mid x)}{f(y \mid x)} \right]$$
$$= \mathop{\mathbb{E}}_{x,y\sim\Pr_{\mathcal{D}}} \left[ \log \frac{1}{f(y \mid x)} - \log \frac{1}{\Pr_{\mathcal{D}}(y \mid x)} \right]$$
$$= \frac{1}{|\mathcal{D}|} \sum_{(x,y)\in\mathcal{D}} \left[ \log \frac{1}{f(y \mid x)} \right] \quad - \mathrm{H}_{\Pr_{\mathcal{D}}}(Y \mid X)$$

□

**Proposition 6.** *Consider functions $f, h : X \to Y$ from inputs to labels, where $h$ is a predictor and $f$ generates the true labels. The inconsistency of believing $f$ and $h$ (with any confidences), and a distribution $D(X)$ with confidence $\beta$, is $\beta$ times the log accuracy of $h$. That is,*

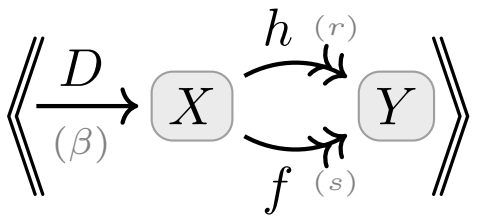

$$= -\beta \log \Pr_{x\sim D}(f(x) = h(x)) = \beta\, \mathrm{I}_D[f = h]. \tag{3}$$

*Proof.* Becuase $f$ is deterministic, for every $x$ in the support of a joint distribution $\mu$ with finite score, we must have $\mu(Y \mid x) = \delta_f$, since if $\mu$ were to place any non-zero mass $\mu(x, y) = \epsilon > 0$ on a pont $(x, y)$ with

$y \neq f(x)$ results in an infinite contribution to the KL divergence

$$\boldsymbol{D}(\mu(Y \mid x) \,\|\, \delta_{f(x)}) = \mathop{\mathbb{E}}_{x,y\sim\mu} \log \frac{\mu(y \mid x)}{\delta_{f(x)}} \geq \mu(y,x) \log \frac{\mu(x,y)}{\mu(x)\cdot\delta_{f(x)}(y)} = \epsilon \log \frac{\epsilon}{0} = \infty.$$

The same holds for $h$. Therefore, for any $\mu$ with a finite score, and $x$ with $\mu(x) > 0$, we have $\delta_{f(x)} = \mu(Y \mid x) = \delta_{h(x)}$, meaning that we need only consider $\mu$ whose support is a subset of those points on which $f$ and $h$ agree. On all such points, the contribution to the score from the edges associated to $f$ and $h$ will be zero, since $\mu$ matches the conditional marginals exactly, and the total incompatibility of such a distribution $\mu$ is equal to the relative entropy $\boldsymbol{D}(\mu \,\|\, D)$, scaled by the confidence $\beta$ of the empirical distribution $D$.

So, among those distributions $\mu(X)$ supported on an event $E \subset \mathcal{V}(X)$, which minimizes is the relative entropy of $\boldsymbol{D}(\mu \,\|\, D)$? It is well known that the conditional distribution $D \mid E \propto \delta_E(X)D(X) = \frac{1}{D(E)}\delta_E(X)D(X)$ satisfies this property uniquely (see, for instance, Fagin et al. 2003). Let $f\!=\!h$ denote the event that $f$ and $h$ agree. Then we calculate

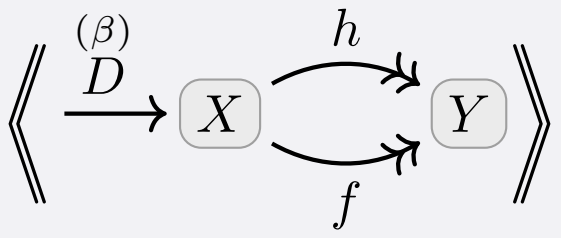

$$\begin{aligned}
&= \inf_{\substack{\mu(X) \text{ s.t.}\\ \mathrm{supp}(\mu)\subseteq[f=h]}} \beta \boldsymbol{D}\Big(\mu(X) \,\Big\|\, D(X)\Big) \\
&= \beta \boldsymbol{D}\Big(D \mid [f\!=\!h] \,\Big\|\, D\Big) \\
&= \beta \mathop{\mathbb{E}}_{D|f=h} \log \frac{\delta_{f=h}(X)D(X)}{D(f\!=\!h)\cdot D(X)} \\
&= \beta \mathop{\mathbb{E}}_{D|f=h} \log \frac{1}{D(f\!=\!h)} && \left[\begin{array}{c}\text{since } \delta_{f=h}(x) = 1 \text{ for all } x \text{ that}\\ \text{contribute to the expectation}\end{array}\right] \\
&= -\beta \log D(f=h) && \left[\ \text{since } D(f=h) \text{ is a constant}\ \right] \\
&= -\beta \log\Big(\mathrm{accuracy}_{f,D}(h)\Big) \\
&= \beta\, \mathrm{I}_D[f=h].
\end{aligned}$$

□

**Proposition 16.** *Consider a PDG containing two (distinct) conditional Gaussian distributions on a variable $Y$, whose parameters can both depend on a variable $X$. Its inconsistency takes the form*

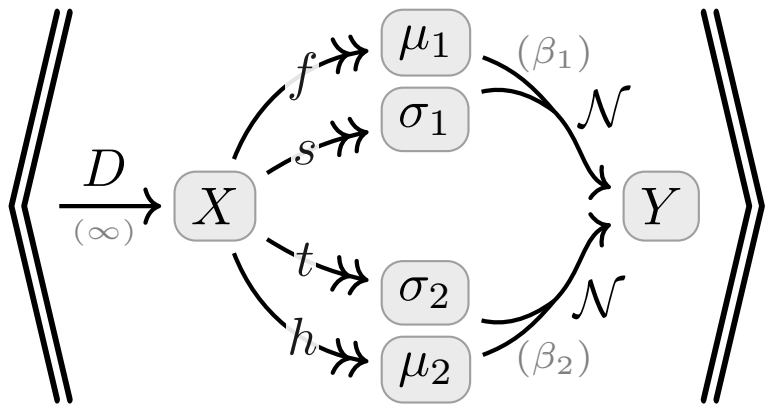

$$= \mathop{\mathbb{E}}_{D}\left[(\beta_1\!+\!\beta_2)\log\frac{\mathrm{QM}_{\hat\beta}(\sigma_1,\sigma_2)}{\mathrm{GM}_{\hat\beta}(\sigma_1,\sigma_2)} + \frac{1}{2}\frac{\beta_1\beta_2}{\beta_1+\beta_2}\left(\frac{\mu_1-\mu_2}{\mathrm{QM}_{\hat\beta}(\sigma_1,\sigma_2)}\right)^2\right] \tag{7}$$

$$= \frac{1}{2}\mathop{\mathbb{E}}_{x\sim D}\left[\frac{\beta_1\beta_2}{2}\frac{\big(f(x)-h(x)\big)^2}{\beta_2 s(x)^2+\beta_1 t(x)^2} + \frac{\beta_1+\beta_2}{2}\log\frac{\beta_2 s(x)^2+\beta_1 t(x)^2}{\beta_1+\beta_2} \begin{array}{l}-\beta_2\log s(x)\\ -\beta_1 \log t(x)\end{array}\right]$$

*where $\hat\beta = (\frac{\beta_2}{\beta_1+\beta_2}, \frac{\beta_1}{\beta_1+\beta_2})$ represents the normalized and reversed vector of conficences $\beta = (\beta_1,\beta_2)$ for the two distributions, and $\mu_1 = f(X)$, $\mu_2 = g(X)$, $\sigma_1 = s(X)$, $\sigma_2 = t(X)$ are random variables over $X$.*

*Proof.* Let $\mathcal{m}$ denote the PDG in question. Since $D$ has high confidence, we know any joint distribution

$\mu$ with a finite score must have $\mu(X) = D(X)$. Thus,

$$\begin{aligned}
\langle\!\langle \mathcal{M} \rangle\!\rangle &= \inf_{\mu} \mathop{\mathbb{E}}_{x\sim D} \mathop{\mathbb{E}}_{y\sim\mu|x} \left[ \beta_1 \log \frac{\mu(y \mid x)}{\mathcal{N}(y \mid f(x), s(x))} + \beta_2 \log \frac{\mu(y \mid x)}{\mathcal{N}(y \mid h(x), t(x))} \right] \\
&= \inf_{\mu} \mathop{\mathbb{E}}_{x\sim D} \mathop{\mathbb{E}}_{y\sim\mu|x} \left[ \beta_1 \log \frac{\mu(y \mid x)}{\frac{1}{s(x)\sqrt{2\pi}} \exp\left(-\frac{1}{2}\left(\frac{y-f(x)}{s(x)}\right)^2\right)} + \beta_2 \log \frac{\mu(y \mid x)}{\frac{1}{t(x)\sqrt{2\pi}} \exp\left(-\frac{1}{2}\left(\frac{y-h(x)}{t(x)}\right)^2\right)} \right] \\
&= \inf_{\mu} \mathop{\mathbb{E}}_{x\sim D} \mathop{\mathbb{E}}_{y\sim\mu|x} \left[ \log \mu(y \mid x)^{\beta_1+\beta_2} \begin{array}{ll} +\frac{\beta_1}{2}\left(\frac{y-f(x)}{s(x)}\right)^2 & +\frac{\beta_2}{2}\left(\frac{y-h(x)}{t(x)}\right)^2 \\ +\beta_1 \log(s(x)\sqrt{2\pi}) & +\beta_2 \log(t(x)\sqrt{2\pi}) \end{array} \right]. \qquad (8)
\end{aligned}$$

At this point, we would like make use of the fact that the sum of two parabolas is itself a parabola, so as to combine the two terms on the top right of the previous equation. Concretely, we claim (Claim 1, whose proof is at the end of the present one), that if we define

$$g(x) := \frac{\beta_1 t(x)^2 f(x) + \beta_2 s(x)^2 h(x)}{\beta_1 t(x)^2 + \beta_2 s(x)^2} \quad \text{and} \quad \tilde\sigma(x) := \frac{s(x)t(x)}{\sqrt{\beta_1 t(x)^2 + \beta_2 s(x)^2}},$$

then

$$\frac{\beta_1}{s(x)^2}(y-f)^2 + \frac{\beta_2}{t(x)^2}(y-h)^2 = \left(\frac{y-g}{\tilde\sigma}\right)^2 + \frac{\beta_1\beta_2}{\beta_1 t(x)^2 + \beta_2 s(x)^2}(f-h)^2.$$

Applying this to (8) leaves us with:

$$\langle\!\langle \mathcal{M} \rangle\!\rangle = \inf_{\mu} \mathop{\mathbb{E}}_{x\sim D} \mathop{\mathbb{E}}_{y\sim\mu|x} \left[ \log \mu(y \mid x)^{\beta_1+\beta_2} \begin{array}{ll} +\frac{1}{2\,\tilde\sigma(x)^2}(y-g(x))^2 & +\frac{1}{2}\frac{\beta_1\beta_2}{\beta_1 t(x)^2+\beta_2 s(x)^2}(f(x)-h(x))^2 \\ +\beta_1 \log(s(x)\sqrt{2\pi}) & +\beta_2 \log(t(x)\sqrt{2\pi}) \end{array} \right]$$

Pulling the term on the top right, which does not depend on $Y$, out of the expectation, and folding the rest of the terms back inside the logarithm (which in particular means first replacing the top middle term $\varphi$ by $-\log(\exp(-\varphi))$), we obtain:

$$\langle\!\langle \mathcal{M} \rangle\!\rangle = \mathop{\mathbb{E}}_{x\sim D} \left[ \begin{array}{l} \inf\limits_{\mu(Y)} \mathop{\mathbb{E}}\limits_{y\sim\mu} \left[ \log \mu(y)^{\beta_1+\beta_2} - \log\left( \frac{1}{\sqrt{2\pi}^{\beta_1+\beta_2} s(x)^{\beta_1} t(x)^{\beta_2}} \exp\left\{-\frac{1}{2}\left(\frac{y-g(x)}{\tilde\sigma(x)}\right)^2\right\}\right)\right] \\ \quad + \frac{1}{2}\frac{\beta_1\beta_2}{\beta_1 t(x)^2 + \beta_2 s(x)^2}\Big(f(x)-h(x)\Big)^2 \end{array} \right].$$

To simplify the presentation, let $\psi$ be the term on the top right, and $\xi$ be the term on the bottom. More explicitly, define

$$\psi(x,y) := \frac{1}{2}\frac{1}{\sqrt{2\pi}^{\beta_1+\beta_2} s(x)^{\beta_1} t(x)^{\beta_2}} \exp\left\{-\frac{1}{2}\left(\frac{y-g(x)}{\tilde\sigma(x)}\right)^2\right\}, \text{ and } \xi(x) := \frac{\beta_1\beta_2}{\beta_1 t(x)^2 + \beta_2 s(x)^2}\Big(f(x)-h(x)\Big)^2,$$

which lets us write the previous expression for $\langle\!\langle \mathcal{M} \rangle\!\rangle$ as

$$\langle\!\langle \mathcal{M} \rangle\!\rangle = \mathop{\mathbb{E}}_{x\sim D}\left[ \inf_{\mu(Y)} \mathop{\mathbb{E}}_{y\sim\mu} \Big[\log \mu(y)^{\beta_1+\beta_2} - \log\psi(x,y)\Big] + \xi(x) \right]. \qquad (9)$$

Also, let $\hat\beta_1 := \frac{\beta_1}{\beta_1+\beta_2}$, and $\hat\beta_2 := \frac{\beta_2}{\beta_1+\beta_2}$. For reasons that will soon become clear, we are actually interested in $\psi^{\frac{1}{\beta_1+\beta_2}}$, which we compute as

$$\begin{aligned}
\psi(x,y)^{\frac{1}{\beta_1+\beta_2}} &= (2\pi)^{-\frac{1}{2}}\, s(x)^{\left(\frac{-\beta_1}{\beta_1+\beta_2}\right)} t(x)^{\left(\frac{-\beta_2}{\beta_1+\beta_2}\right)} \exp\left\{-\frac{1}{2}\left(\frac{y-g(x)}{\tilde\sigma(x)}\right)^2\right\}^{\frac{1}{\beta_1+\beta_2}} \\
&= \frac{1}{\sqrt{2\pi}\, s(x)^{\hat\beta_1} t(x)^{\hat\beta_2}} \exp\left\{\frac{-1}{2(\beta_1+\beta_2)}\left(\frac{y-g(x)}{\tilde\sigma(x)}\right)^2\right\}.
\end{aligned}$$

Recall that the Gaussian densitity $\mathcal{N}(y|\, g(x),\, \tilde\sigma(x)\sqrt{\beta_1+\beta_2})$ of mean $g(x)$ and variance $\tilde\sigma(x)^2(\beta_1+\beta_2)$ is given by

$$\mathcal{N}\left(y \,\middle|\, g(x),\, \tilde\sigma(x)\sqrt{\beta_1+\beta_2}\right) = \frac{1}{\sqrt{2\pi}\;\tilde\sigma(x)\sqrt{\beta_1+\beta_2}} \exp\left\{\frac{-1}{2(\beta_1+\beta_2)}\left(\frac{y-g(x)}{\tilde\sigma(x)}\right)^2\right\},$$

which is quite similar, and has an identical dependence on $y$. To facilitate converting one to the other, we explicitly compute the ratio:

$$\begin{aligned}
\frac{\psi(x,y)^{\frac{1}{\beta_1+\beta_2}}}{\mathcal{N}\left(y \,\middle|\, g(x),\, \tilde\sigma(x)\sqrt{\beta_1+\beta_2}\right)} &= \frac{\tilde\sigma\sqrt{2\pi(\beta_1+\beta_2)}}{\sqrt{2\pi}\; s(x)^{\hat\beta_1}t(x)^{\hat\beta_2}} = \frac{\tilde\sigma\sqrt{\beta_1+\beta_2}}{s(x)^{\hat\beta_1}t(x)^{\hat\beta_2}} \\
&= \left(\frac{s(x)t(x)}{\sqrt{\beta_1 t(x)^2+\beta_2 s(x)^2}}\right)\frac{\sqrt{\beta_1+\beta_2}}{s(x)^{\hat\beta_1}t(x)^{\hat\beta_2}} && [\text{expand defn of } \tilde\sigma(x)] \\
&= s(x)^{1-\hat\beta_1}\, t(x)^{1-\hat\beta_2}\sqrt{\frac{\beta_1+\beta_2}{\beta_1\, t(x)^2+\beta_2\, s(x)^2}} \\
&= s(x)^{1-\hat\beta_1}\, t(x)^{1-\hat\beta_2}\sqrt{\frac{1}{\hat\beta_1\, t(x)^2+\hat\beta_2\, s(x)^2}} && [\text{defn of } \hat\beta_1,\hat\beta_2] \\
&= \frac{s(x)^{\hat\beta_2}\, t(x)^{\hat\beta_1}}{\sqrt{\hat\beta_1\, t(x)^2+\hat\beta_2\, s(x)^2}} && [\text{since } \hat\beta_1+\hat\beta_2=1]
\end{aligned}$$

Now, picking up from where we left off in (9), we have

$$\begin{aligned}
⟪m⟫ &= \mathop{\mathbb{E}}_{x\sim D}\left[\inf_{\mu(Y)}\mathop{\mathbb{E}}_{y\sim\mu}\left[\log\mu(y)^{\beta_1+\beta_2}-\log\psi(x,y)\right]+\xi(x)\right] \\
&= \mathop{\mathbb{E}}_{x\sim D}\left[\inf_{\mu(Y)}\mathop{\mathbb{E}}_{y\sim\mu}\left[\log\frac{\mu(y)^{\beta_1+\beta_2}}{\psi(x,y)^{\frac{\beta_1+\beta_2}{\beta_1+\beta_2}}}\right]+\xi(x)\right] \\
&= \mathop{\mathbb{E}}_{x\sim D}\left[\inf_{\mu(Y)}\mathop{\mathbb{E}}_{y\sim\mu}\left[(\beta_1+\beta_2)\log\frac{\mu(y)}{\psi(x,y)^{\frac{1}{\beta_1+\beta_2}}}\right]+\xi(x)\right] \\
&= \mathop{\mathbb{E}}_{x\sim D}\left[\inf_{\mu(Y)}\mathop{\mathbb{E}}_{y\sim\mu}\left[(\beta_1+\beta_2)\log\frac{\mu(y)}{\mathcal{N}\left(y \,\middle|\, g(x),\, \tilde\sigma(x)\sqrt{\beta_1+\beta_2}\right)\frac{s(x)^{\hat\beta_2}\, t(x)^{\hat\beta_1}}{\sqrt{\hat\beta_1\, t(x)^2+\hat\beta_2\, s(x)^2}}}\right]+\xi(x)\right] \\
&= \mathop{\mathbb{E}}_{x\sim D}\left[\inf_{\mu(Y)}\mathop{\mathbb{E}}_{y\sim\mu}\left[(\beta_1+\beta_2)\log\frac{\mu(y)}{\mathcal{N}\left(y \,\middle|\, g(x),\, \tilde\sigma(x)\sqrt{\beta_1+\beta_2}\right)}\right]+(\beta_1+\beta_2)\log\frac{\sqrt{\hat\beta_1\, t(x)^2+\hat\beta_2\, s(x)^2}}{s(x)^{\hat\beta_2}\, t(x)^{\hat\beta_1}}+\xi(x)\right]
\end{aligned}$$

but now the entire left term is the infemum of a KL divergence, which is non-negative and equal to zero iff $\mu(y)=\mathcal{N}(y|g(x),\tilde\sigma(x)\sqrt{\beta_1+\beta_2})$. So the infemum on the left is equal to zero.

$$⟪m⟫ = \mathop{\mathbb{E}}_{x\sim D}\left[(\beta_1+\beta_2)\log\frac{\sqrt{\hat\beta_1\, t(x)^2+\hat\beta_2\, s(x)^2}}{s(x)^{\hat\beta_2}\, t(x)^{\hat\beta_1}}+\xi(x)\right] \tag{10}$$

$$\begin{aligned}
&= \mathop{\mathbb{E}}_{x\sim D}\left[(\beta_1+\beta_2)\log\sqrt{\hat\beta_1\, t(x)^2+\hat\beta_2\, s(x)^2}-(\beta_1+\beta_2)\log\left(s(x)^{\hat\beta_2}\, t(x)^{\hat\beta_1}\right)+\xi(x)\right] \\
&= \mathop{\mathbb{E}}_{x\sim D}\left[(\beta_1+\beta_2)\log\sqrt{\hat\beta_1\, t(x)^2+\hat\beta_2\, s(x)^2}\;\begin{matrix}-\beta_2\log s(x)\\-\beta_1\log t(x)\end{matrix}\;+\xi(x)\right] \\
&= \mathop{\mathbb{E}}_{x\sim D}\left[(\beta_1+\beta_2)\log\sqrt{\frac{\beta_1\, t(x)^2+\beta_2\, s(x)^2}{\beta_1+\beta_2}}\;\begin{matrix}-\beta_2\log s(x)\\-\beta_1\log t(x)\end{matrix}\;+\frac{1}{2}\frac{\beta_1\beta_2}{\beta_1 t(x)^2+\beta_2 s(x)^2}\Big(f(x)-h(x)\Big)^2\right]
\end{aligned} \tag{11}$$

Whew! Pulling the square root of the logarithm proves complex second half of the proposition. Now, we massage it into into a (slightly) more readable form.

To start, write $\sigma_1$ (the random variable) in place of $s(x)$ and $\sigma_2$ in place of $t(x)$. Let $\hat\beta$ without the subscript denote the vector $(\hat\beta_2, \hat\beta_1) = (\frac{\beta_2}{\beta_1+\beta_2}, \frac{\beta_1}{\beta_1+\beta_2})$, which we will use for weighted means. The $\hat\beta$-weighted arithmetic, geometric ($p = 0$), and quadratic ($p = 2$) means of $\sigma_1$ and $\sigma_2$ are:

$$\mathrm{GM}_{\hat\beta}(\sigma_1, \sigma_2) = (\sigma_1)^{\hat\beta_2}(\sigma_2)^{\hat\beta_1} \qquad \text{and} \qquad \mathrm{QM}_{\hat\beta}(\sigma_1, \sigma_2) = \sqrt{\hat\beta_2\sigma_1^2 + \hat\beta_1\sigma_2^2}.$$

So, now we can write $\xi(x)$ as

$$\begin{aligned}
\frac{1}{2}\frac{\beta_1\beta_2}{\beta_1 t(x)^2 + \beta_2 s(x)^2}\Big(f(x) - h(x)\Big)^2 &= \frac{1}{2}\frac{\beta_1\beta_2}{\beta_1+\beta_2}\frac{\beta_1+\beta_2}{\beta_1 t(x)^2 + \beta_2 s(x)^2}\Big(f(x) - h(x)\Big)^2 \\
&= \frac{1}{2}\frac{\beta_1\beta_2}{\beta_1+\beta_2}\left(\frac{1}{\mathrm{QM}_{\hat\beta}(\sigma_1,\sigma_2)}\right)^2\Big(f(x) - h(x)\Big)^2 \\
&= \frac{1}{2}\frac{\beta_1\beta_2}{\beta_1+\beta_2}\left(\frac{\mu_1-\mu_2}{\mathrm{QM}_{\hat\beta}(\sigma_1,\sigma_2)}\right)^2;
\end{aligned}$$

in the last step, we have replaced $f(x)$ and $g(x)$ with their respective random variables $\mu_1$ and $\mu_2$. As a result, (10) can be written as

$$\langle\!\langle m \rangle\!\rangle = \mathop{\mathbb{E}}_{D}\left[(\beta_1+\beta_2)\log\frac{\mathrm{QM}_{\hat\beta}(\sigma_1,\sigma_2)}{\mathrm{GM}_{\hat\beta}(\sigma_1,\sigma_2)} + \frac{1}{2}\frac{\beta_1\beta_2}{\beta_1+\beta_2}\left(\frac{\mu_1-\mu_2}{\mathrm{QM}_{\hat\beta}(\sigma_1,\sigma_2)}\right)^2\right]$$

. . . which is perhaps more comprehensible, and proves the first half of our proposition. □

**Claim 1.** *The sum of two functions that are unshifted parabolas as functions of $y$ (i.e., both functions are of of the form $k(y-a)^2$), is itself a (possibly shifted) parabola of $y$ (and of the form $k'(y-a')+b'$). More concretely, and adapted to our usage above, the following algebraic relation holds:*

$$\frac{\beta_1}{\sigma_1^2}(y-f)^2 + \frac{\beta_2}{\sigma_2^2}(y-h)^2 = \left(\frac{y-g}{\tilde\sigma}\right)^2 + \frac{\beta_1\beta_2}{\beta_1\sigma_2^2+\beta_2\sigma_1^2}(f-h)^2,$$

*where*

$$g := \frac{\beta_1\sigma_2^2 f + \beta_2\sigma_1^2 h}{\beta_1\sigma_2^2+\beta_2\sigma_1^2} \quad \text{and} \quad \tilde\sigma := \left(\frac{\beta_1}{\sigma_1^2}+\frac{\beta_2}{\sigma_2^2}\right)^{-1/2} = \frac{\sigma_1\sigma_2}{\sqrt{\beta_1\sigma_2^2+\beta_2\sigma_1^2}}.$$

*Proof.* Expand terms and complete the square. Starting from the left hand side, we have

$$\begin{aligned}
&\frac{\beta_1}{\sigma_1^2}(y-f)^2 + \frac{\beta_2}{\sigma_2^2}(y-h)^2 \\
&= \frac{\beta_1}{\sigma_1^2}(y^2 - 2yf + f^2) + \frac{\beta_2}{\sigma_2^2}(y^2 - 2yh + h^2) \\
&= \left(\frac{\beta_1}{\sigma_1^2}+\frac{\beta_2}{\sigma_2^2}\right)y^2 - 2\left(\frac{\beta_1 f}{\sigma_1^2}+\frac{\beta_2 h}{\sigma_2^2}\right)y + \left(\frac{\beta_1 f^2}{\sigma_1^2}+\frac{\beta_2 h^2}{\sigma_2^2}\right) \\
&= \left(\frac{\beta_1}{\sigma_1^2}+\frac{\beta_2}{\sigma_2^2}\right)y^2 - 2\left(\frac{\beta_1 f}{\sigma_1^2}+\frac{\beta_2 h}{\sigma_2^2}\right)y + \left(\frac{\beta_1 f^2}{\sigma_1^2}+\frac{\beta_2 h^2}{\sigma_2^2} - \frac{\beta_1\beta_2(f-h)^2}{\beta_1\sigma_2^2+\beta_2\sigma_1^2}\right) + \frac{\beta_1\beta_2(f-h)^2}{\beta_1\sigma_2^2+\beta_2\sigma_1^2} \qquad (12)
\end{aligned}$$

where in the last step we added and removed the same term (i.e., the completion of the square, although it is probably still unclear why this quantity will do that). The third parenthesized quantity needs the most work. Isolating it and getting a common denominator gives us:

$$\begin{aligned}
&\frac{\beta_1 f^2}{\sigma_1^2} + \frac{\beta_2 h^2}{\sigma_2^2} - \frac{\beta_1\beta_2(f-h)^2}{\beta_1\sigma_2^2+\beta_2\sigma_1^2} \\
&= \frac{\beta_1 f^2(\beta_1\sigma_2^2+\beta_2\sigma_1^2)\sigma_2^2}{\sigma_1^2(\beta_1\sigma_2^2+\beta_2\sigma_1^2)\sigma_2^2} + \frac{\beta_2 h^2(\beta_1\sigma_2^2+\beta_2\sigma_1^2)\sigma_1^2}{\sigma_2^2(\beta_1\sigma_2^2+\beta_2\sigma_1^2)\sigma_1^2} - \frac{\beta_1\beta_2(f^2-2fh+h^2)(\sigma_1^2\sigma_2^2)}{(\beta_1\sigma_2^2+\beta_2\sigma_1^2)(\sigma_1^2\sigma_2^2)} \\
&= \frac{\beta_1^2\sigma_2^4 f^2 + \cancel{\beta_1\beta_2\sigma_2^2\sigma_1^2 f^2} + \cancel{\beta_1\beta_2\sigma_1^2\sigma_2^2 h^2} + \beta_2^2\sigma_1^4 h^2 - \cancel{\beta_1\beta_2\sigma_1^2\sigma_2^2 f^2} + 2\beta_1\beta_2\sigma_1^2\sigma_2^2 fh - \cancel{\beta_1\beta_2\sigma_1^2\sigma_2^2 h^2}}{(\beta_1\sigma_2^2+\beta_2\sigma_1^2)(\sigma_1^2\sigma_2^2)} \\
&= \frac{\beta_1^2\sigma_2^4 f^2 + \beta_2^2\sigma_1^4 h^2 + 2\beta_1\beta_2\sigma_1^2\sigma_2^2 fh}{(\beta_1\sigma_2^2+\beta_2\sigma_1^2)(\sigma_1^2\sigma_2^2)}.
\end{aligned}$$

Substituting this expression into the third term of (12), while simultaneously computing common denominators for the first and second terms, yields

$$\left(\frac{\beta_1\sigma_2^2+\beta_2\sigma_1^2}{\sigma_1^2\sigma_2^2}\right)y^2 - 2\left(\frac{\beta_1\sigma_2^2 f+\beta_2\sigma_1^2 h}{\sigma_1^2\sigma_2^2}\right)y + \frac{\beta_1^2\sigma_2^4 f^2 + \beta_2^2\sigma_1^4 h^2 + 2\beta_1\beta_2\sigma_1^2\sigma_2^2 fh}{(\beta_1\sigma_2^2+\beta_2\sigma_1^2)(\sigma_1^2\sigma_2^2)} + \frac{\beta_1\beta_2(f-h)^2}{\beta_1\sigma_2^2+\beta_2\sigma_1^2}. \tag{13}$$

On the other hand, using the definitions of $g$ and $\tilde{\sigma}$, we compute:

$$\begin{aligned}
\left(\frac{y-g}{\tilde{\sigma}}\right)^2 &= \left(\frac{\beta_1\sigma_2^2+\beta_2\sigma_1^2}{\sigma_1^2\sigma_2^2}\right)\left(y - \frac{\beta_1\sigma_2^2 f+\beta_2\sigma_1^2 h}{\beta_1\sigma_2^2+\beta_2\sigma_1^2}\right)^2 \\
&= \left(\frac{\beta_1\sigma_2^2+\beta_2\sigma_1^2}{\sigma_1^2\sigma_2^2}\right)\left(y^2 - 2y\frac{\beta_1\sigma_2^2 f+\beta_2\sigma_1^2 h}{\beta_1\sigma_2^2+\beta_2\sigma_1^2} + \frac{\beta_1^2\sigma_2^4 f^2 + \beta_2^2\sigma_1^4 h^2 + 2\beta_1\beta_2\sigma_2^2\sigma_1^2\, fh}{(\beta_1\sigma_2^2+\beta_2\sigma_1^2)^2}\right) \\
&= \left(\frac{\beta_1\sigma_2^2+\beta_2\sigma_1^2}{\sigma_1^2\sigma_2^2}\right)y^2 - 2\left(\frac{\beta_1\sigma_2^2 f+\beta_2\sigma_1^2 h}{\sigma_1^2\sigma_2^2}\right)y + \frac{\beta_1^2\sigma_2^4 f^2 + \beta_2^2\sigma_1^4 h^2 + 2\beta_1\beta_2\sigma_2^2\sigma_1^2\, fh}{(\beta_1\sigma_2^2+\beta_2\sigma_1^2)(\sigma_1^2\sigma_2^2)}
\end{aligned}$$

. . . which is precisely the first 3 terms of (13). Putting it all together, we have shown that

$$\frac{\beta_1}{\sigma_1^2}(y-f)^2 + \frac{\beta_2}{\sigma_2^2}(y-h)^2 = \left(\frac{y-g}{\tilde{\sigma}}\right)^2 + \frac{\beta_1\beta_2(f-h)^2}{\beta_1\sigma_2^2+\beta_2\sigma_1^2}$$

as desired. □

**Proposition 7.**

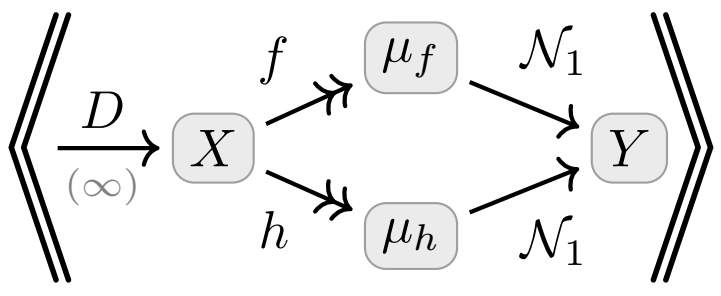

$$= \frac{1}{2}\,\mathbb{E}_D\big|f(X)-h(X)\big|^2 =: \mathrm{MSE}_D(f,h)\,,$$

*where $\mathcal{N}_1(Y|\,\mu)$ is a unit Gaussian on $Y$ with mean $\mu$.*

*Proof.* An immediate corolary of Proposition 16; simply set $s(x) = t(x) = \beta_1 = \beta_2 = 1$ □

**Lemma 10.** *The inconsistency $\boldsymbol{D}^{\mathrm{PDG}}_{(r,s)}(p\|q)$ of a PDG comprising $p(X)$ with confidence $r$ and $q(X)$ with confidence $s$ is given in closed form by*

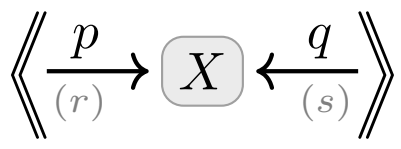

$$= -(r+s)\log\sum_x \Big(p(x)^r q(x)^s\Big)^{\frac{1}{r+s}}.$$

*Proof.*

$$\left\langle\!\!\left\langle \xrightarrow[(\beta:r)]{p} \boxed{X} \xleftarrow[(\beta:s)]{q} \right\rangle\!\!\right\rangle = \inf_{\mu} \mathop{\mathbb{E}}_{\mu} \log \frac{\mu(x)^{r+s}}{p(x)^r q(x)^s}$$

$$= (r+s) \inf_{\mu} \mathop{\mathbb{E}}_{\mu} \left[ \log \frac{\mu(x)}{(p(x)^r q(x)^s)^{\frac{1}{r+s}}} \cdot \frac{Z}{Z} \right]$$

$$= \inf_{\mu} (r+s) \boldsymbol{D}\left( \mu \,\middle\|\, \frac{1}{Z} p^{\frac{r}{r+s}} q^{\frac{s}{r+s}} \right) - (r+s) \log Z$$

where $Z := (\sum_x p(x)^r q(x)^s)^{\frac{1}{r+s}}$ is the constant required to normalize the denominator as a distribution. The first term is now a relative entropy, and the only usage of $\mu$. $\boldsymbol{D}(\mu \parallel \cdots)$ achives its minimum of zero when $\mu$ is the second distribution, so our formula becomes

$$= -(r+s) \log Z$$

$$= -(r+s) \log \sum_x \Big( p(x)^r q(x)^s \Big)^{\frac{1}{r+s}} \qquad \text{as promised.}$$

□

**Proposition 8.** *Suppose you have a parameterized model $p(Y|\Theta)$, a prior $q(\Theta)$, and a trusted distribution $D(Y)$. The inconsistency of also believing $\Theta = \theta$ is the cross entropy loss, plus the regularizer: $\log \frac{1}{q(\theta)}$ times your confidence in $q$. That is,*

$$\left\langle\!\!\left\langle \xrightarrow[(\beta)]{q} \boxed{\Theta} \xleftarrow{\theta},\; \boxed{\Theta} \xrightarrow{p} \boxed{Y} \xleftarrow[(\infty)]{D} \right\rangle\!\!\right\rangle = \mathop{\mathbb{E}}_{y\sim D} \log \frac{1}{p(y \,|\, \theta)} + \beta \log \frac{1}{q(\theta)} - \mathrm{H}(D) \tag{4}$$

*Proof.* This is another case where there's only one joint distribution $\mu(\Theta, Y)$ that gets a finite score. We must have $\mu(Y) = D(Y)$ since $D$ has infinite confidence, which uniquely extends to the distribution $\mu(\Theta, Y) = D(Y)\delta_\theta(\Theta)$ for which deterministically sets $\Theta = \theta$.

The cpds corresponding to the edges labeled $\theta$ and $D$, then, are satisfied by this $\mu$ and contribute nothing to the score. So the two relevant edges that contribute incompatibility with this distribution are $p$ and $q$. Letting $\mathcal{m}$ denote the PDG in question, we compute:

$$\langle\!\langle \mathcal{m} \rangle\!\rangle = \mathop{\mathbb{E}}_{\mu} \left[ \log \frac{\mu(Y|\Theta)}{p(Y|\Theta)} + \beta \log \frac{\mu(\Theta)}{q(\Theta)} \right]$$

$$= \mathop{\mathbb{E}}_{y\sim D} \left[ \log \frac{D(y)}{p(y|\theta)} + \beta \log \frac{1}{q(\theta)} \right]$$

$$= \mathop{\mathbb{E}}_{y\sim D} \left[ \log \frac{1}{p(y|\theta)} + \beta \log \frac{1}{q(\theta)} + \log D(y) \right]$$

$$= \mathop{\mathbb{E}}_{y\sim D} \left[ \log \frac{1}{p(y|\theta)} \right] + \beta \log \frac{1}{q(\theta)} - \mathrm{H}(D)$$

as desired. □

**Proposition 11.** *The negative ELBO of $x$ is the inconsistency of the PDG containing $p$,$q$, and $X{=}x$, with high confidence in $q$. That is,*

$$-\mathrm{ELBO}_{p,q}(x) =$$

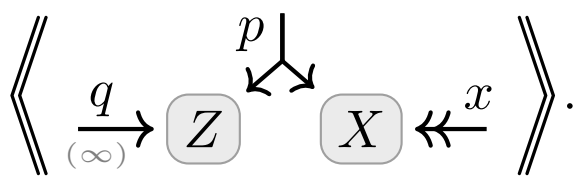

*Proof.* Every distribution that does marginalize to $q(Z)$ or places any mass on $x' \neq x$ will have infinite score. Thus the only distribution that could have a finite score is $\mu(X, Z)$. Thus,

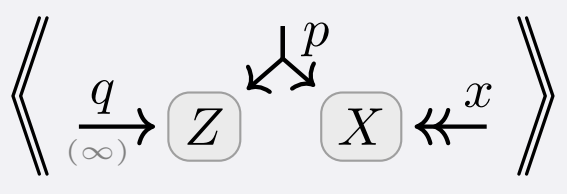

$$= \inf_{\mu}$$

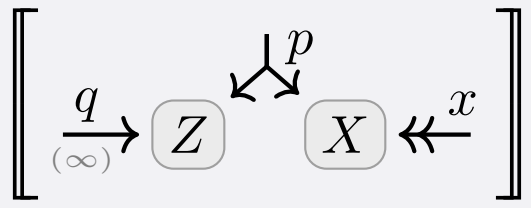

$$(\mu)$$

$$=$$

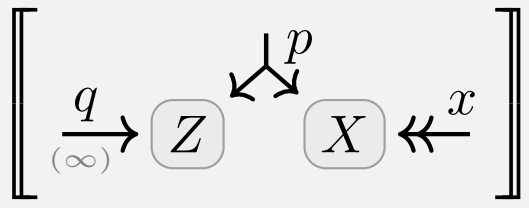

$$(\delta_x(X)q(Z))$$

$$= \mathop{\mathbb{E}}_{\substack{x' \sim \delta_x \\ z \sim q}} \log \frac{\delta_x(x')q(z)}{p(x', z)} = - \mathop{\mathbb{E}}_{z \sim q} \frac{p(x,z)}{q(z)} = -\mathrm{ELBO}_{p,q}(x).$$

□

We proove both Proposition 12 and Proposition 17 at the same time.

**Proposition 12.** *The VAE loss of a sample $x$ is the inconsistency of the PDG comprising the encoder $e$ (with high confidence, as it defines the encoding), decoder $d$, prior $p$, and $x$. That is,*

$$-\mathrm{ELBO}_{p,e,d}(x) =$$

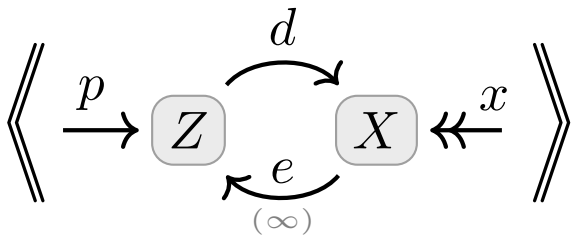

**Proposition 17.** *The following analog of Proposition 12 for a whole dataset $\mathcal{D}$ holds:*

$$- \mathop{\mathbb{E}}_{\Pr_{\mathcal{D}}} \mathrm{ELBO}_{p,e,d}(X) =$$

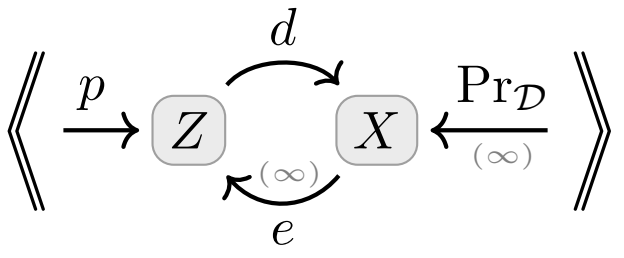

$$+ \mathrm{H}(\Pr_{\mathcal{D}}).$$

*Proof.* The two proofs are similar. For Proposition 12, the optimal distribution must be $\delta_x(X)e(Z \mid X)$, and for Proposition 17, it must be $\Pr_{\mathcal{D}}(X)e(Z \mid X)$, because $e$ and the data both have infinite confidence, so any other distribution gets an infinite score. At the same time, $d$ and $p$ define a joint distribution, so the inconsistency in question becomes

$$\boldsymbol{D}\Big(\delta_x(X)e(Z \mid X) \,\Big\|\, p(Z)d(X \mid Z)\Big) = \mathop{\mathbb{E}}_{z \sim e|x} \left[\log \frac{p(z)d(x \mid z)}{e(z \mid x)}\right] = \mathrm{ELBO}_{p,e,d}(x)$$

in the first, case, and

$$\boldsymbol{D}\Big(\Pr_{\mathcal{D}}(X)e(Z \mid X) \,\Big\|\, p(Z)d(X \mid Z)\Big) = \frac{1}{|\mathcal{D}|} \sum_{x \in \mathcal{D}} \mathop{\mathbb{E}}_{z \sim e|x} \left[\log \frac{p(z)d(x \mid z)}{e(z \mid x)} + \log \frac{1}{\Pr_{\mathcal{D}}(x)}\right]$$
$$= \mathrm{ELBO}_{p,e,d}(x) - \mathrm{H}(\Pr_{\mathcal{D}})$$

in the second. □

Now, we formally state and prove the more general result for $\beta$-VAEs.

**Proposition 18.** *The negative $\beta$-ELBO objective for a prior $p(X)$, encoder $e(Z \mid X)$, decoder $d(X \mid Z)$, at a sample $x$, is equal to the inconsistency of the corresponding PDG, where the prior has confidence equal to $\beta$. That is,*

$$-\beta\text{-ELBO}_{p,e,d}(x) = \left\langle\!\!\left\langle \xrightarrow[(\beta)]{p} \boxed{Z} \underset{e\;(\infty)}{\overset{d}{\rightleftarrows}} \boxed{X} \overset{x}{\twoheadleftarrow} \right\rangle\!\!\right\rangle$$

*Proof.*

$$\left\langle\!\!\left\langle \xrightarrow[(\beta)]{p} \boxed{Z} \underset{e\;(\infty)}{\overset{d}{\rightleftarrows}} \boxed{X} \overset{x}{\twoheadleftarrow} \right\rangle\!\!\right\rangle = \inf_{\mu} \left[\!\!\left[ \xrightarrow[(\beta)]{p} \boxed{Z} \underset{e\;(\infty)}{\overset{d}{\rightleftarrows}} \boxed{X} \overset{x}{\twoheadleftarrow} \right]\!\!\right](\mu)$$

$$= \inf_{\mu} \mathop{\mathbb{E}}_{\mu(X,Z)} \left[ \beta \log \frac{\mu(Z)}{p(Z)} + \log \frac{\mu(X,Z)}{\mu(Z) d(X \mid Z)} \right]$$

As before, the only candidate for a joint distribution with finite score is $\delta_x(X)e(Z \mid X)$. Note that the marginal on $Z$ for this distribution is itself, since $\int_x \delta_x(X) e(Z \mid X) \, \mathrm{d}x = e(Z \mid x)$. Thus, our equation becomes

$$\begin{aligned}
&= \mathop{\mathbb{E}}_{\delta_x(X)e(Z|X)} \left[ \beta \log \frac{e(Z \mid x)}{p(z)} + \log \frac{\delta_x(X) e(Z \mid X)}{e(Z \mid x) d(x \mid Z)} \right] \\
&= \mathop{\mathbb{E}}_{e(Z|x)} \left[ \beta \log \frac{e(Z \mid x)}{p(Z)} + \log \frac{1}{d(x \mid Z)} \right] \\
&= \boldsymbol{D}(e(Z|\,x) \parallel p) + \mathrm{Rec}_{e,d}(x) \\
&= -\beta\text{-ELBO}_{p,e,d}(x).
\end{aligned}$$

□

**Proposition 14.** *For all weighted factor graphs $\Psi$, we have that $\langle\!\langle \mathcal{M}_\Psi \rangle\!\rangle_1 = -\log Z_\Psi$.*

*Proof.* In the main text, we defined $\mathcal{M}_\Psi$ to be the PDG with edges $\{\xrightarrow{J} \mathbf{X}_J\}_{\mathcal{J}}$, cpds $p_J(\mathbf{X}_J) \propto \phi_J(\mathbf{X}_J)$, and weights $\alpha_J, \beta_J := \theta_J$. Let the$(\{x\}) := x$ be a function that extracts the unique element singleton set. It was shown by Richardson and Halpern (2021) (Corolary 4.4.1) that

$$\text{the}[\![\mathcal{M}_\Psi]\!]_1^* = \Pr_{\Phi,\theta}(\mathbf{x}) = \frac{1}{Z_\Psi} \prod_J \phi_J(\mathbf{x}_J)^{\theta_J}.$$

Recall the statement of Prop 4.6 from Richardson and Halpern (2021):

$$[\![\mathcal{M}]\!]_\gamma(\mu) = \mathop{\mathbb{E}}_{\mathbf{w}\sim\mu} \left\{ \sum_{X \xrightarrow{L} Y} \left[ \beta_L \log \frac{1}{\mathbf{p}_L(y^{\mathbf{w}}|x^{\mathbf{w}})} + (\gamma\alpha_L - \beta_L) \log \frac{1}{\mu(y^{\mathbf{w}}|x^{\mathbf{w}})} \right] - \gamma \log \frac{1}{\mu(\mathbf{w})} \right\}, \tag{14}$$

where $x^{\mathbf{w}}$ and $y^{\mathbf{w}}$ are the respective values of the variables $X$ and $Y$ in the world $\mathbf{w}$. Note that if $\gamma = 1$, and $\alpha, \beta$ are both equal to $\theta$ in $\mathcal{M}_\Psi$, the middle term (in purple) is zero. So in our case, since the edges are $\{\xrightarrow{J} \mathbf{X}_J\}$ and $\mathbf{p}_J(\mathbf{X}_J) = \phi_J(\mathbf{X}_J)$, (14) reduces to the standard variational free energy

$$\mathit{VFE}_\Psi(\mu) = \mathop{\mathbb{E}}_{\mu} \left[ \sum_{J \in \mathcal{J}} \theta_J \log \frac{1}{\phi_J(\mathbf{X}_J)} \right] - \mathrm{H}(\mu) \tag{15}$$

$$= \mathop{\mathbb{E}}_{\mu} \langle \boldsymbol{\varphi}, \boldsymbol{\theta} \rangle_{\mathcal{J}} - \mathrm{H}(\mu), \quad \text{where } \varphi_J(\mathbf{X}_J) := \log \frac{1}{\phi_J(\mathbf{X}_J)}.$$

By construction, $\Pr_\Psi$ uniquely minimizes *VFE*. The 1-inconsistency, $\langle\!\langle \mathcal{m}_\Psi \rangle\!\rangle$ is the minimum value attained. We calculate:

$$\begin{aligned}
\langle\!\langle \mathcal{m} \rangle\!\rangle_1 &= \mathit{VFE}_\Psi(\Pr_\Psi) \\
&= \mathop{\mathbb{E}}_{\mathbf{x}\sim\mu} \left\{ \sum_{J\in\mathcal{J}} \left[ \theta_J \log \frac{1}{\phi_J(\mathbf{x}_J)} \right] - \log \frac{1}{\Pr_{\Phi,\theta}(\mathbf{x})} \right\} && \left[ \text{by (15)} \right] \\
&= \mathop{\mathbb{E}}_{\mathbf{x}\sim\mu} \left\{ \sum_{J\in\mathcal{J}} \left[ \theta_J \log \frac{1}{\phi_J(\mathbf{x}_J)} \right] - \log \frac{Z_\Psi}{\prod_{J\in\mathcal{J}} \phi_J(\mathbf{x}_J)^{\theta_j}} \right\} && \left[ \text{definition of } \Pr_\Psi \right] \\
&= \mathop{\mathbb{E}}_{\mathbf{x}\sim\mu} \left\{ \sum_{J} \left[ \theta_J \log \frac{1}{\phi_J(\mathbf{x}_J)} \right] - \sum_{J\in\mathcal{J}} \left[ \theta_J \log \frac{1}{\phi_J(\mathbf{x}_J)} \right] - \log Z_\Psi \right\} \\
&= \mathop{\mathbb{E}}_{\mathbf{x}\sim\mu} [-\log Z_\Psi] \\
&= -\log Z_\Psi && \left[ Z_\Psi \text{ is constant in } \mathbf{x} \right]
\end{aligned}$$

□

**Proposition 15.** $\left\langle\!\!\left\langle \xrightarrow[(\infty)]{p} \boxed{X} \xrightarrow{\hat{c}} \boxed{\mathtt{T}} \twoheadleftarrow^{\mathtt{t}} \right\rangle\!\!\right\rangle = \mathop{\mathbb{E}}_{x\sim p} c(x).$

*Proof.* Since $p$ has high confidence, and $\mathtt{T}$ is always equal to $\mathtt{t}$, the only joint distribution on $(X, \mathtt{T})$ with finite score is $\mu(X, \mathtt{T}) = p(X)\delta_{\mathtt{t}}(\mathtt{T})$. We compute its score directly:

$$\begin{aligned}
\left\langle\!\!\left\langle \xrightarrow[(\infty)]{p} \boxed{X} \xrightarrow{\hat{c}} \boxed{\mathtt{T}} \twoheadleftarrow^{\mathtt{t}} \right\rangle\!\!\right\rangle &= \mathop{\mathbb{E}}_{\mu} \log \frac{\mu(X, \mathtt{T})}{\hat{c}(\mathtt{t} \,|\, X)} = \mathop{\mathbb{E}}_{p} \log \frac{1}{\hat{c}(\mathtt{t} \,|\, X)} = \mathop{\mathbb{E}}_{p} \log \frac{1}{\exp(-c(X))} \\
&= \mathop{\mathbb{E}}_{p} \log \exp(c(X)) = \mathop{\mathbb{E}}_{p} c(X) = \mathop{\mathbb{E}}_{x\sim p} c(x).
\end{aligned}$$

□

## C.1 Additional Proofs for Unnumbered Claims

### C.1.1 Details on the Data Processing Inequality Proof

We now provide more details on the proof of the Data Processing Equality that appeared in Figure 2 of the main text. We repeat it now for convenience, with labeled PDGs $(\mathcal{m}_1, \ldots, \mathcal{m}_5)$ and numbered (in)equalities.

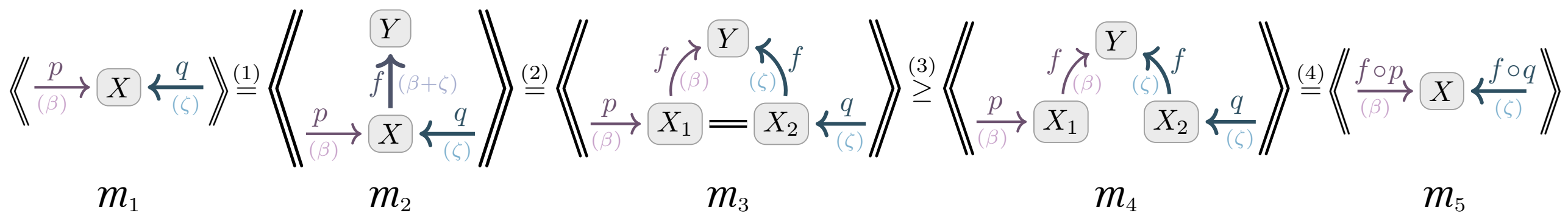

We now enumerate the (in)equalities to prove them.

1. Let $\mu(X)$ denote the (unique) optimal distribution for $\mathcal{m}_1$. Now, the joint distribution $\mu(X, Y) :=$

$\mu(X)f(Y|X)$ has incompatibility with $\mathcal{m}_2$ equal to

$$\begin{aligned} Inc_{\mathcal{m}_2}(\mu(X,Y)) &= \beta \boldsymbol{D}(\mu(X) \parallel p(X)) + \zeta \boldsymbol{D}(\mu(X) \parallel q(X)) + (\beta{+}\zeta) \mathop{\mathbb{E}}_{x\sim\mu} \left[\boldsymbol{D}(\mu(Y|x) \parallel f(Y|x))\right] \\ &= Inc_{\mathcal{m}_1}(\mu(X)) + (\beta{+}\zeta) \mathop{\mathbb{E}}_{x\sim\mu} \boldsymbol{D}(\mu(Y|x) \parallel f(Y|x)) \\ &= \langle\!\langle \mathcal{m}_1 \rangle\!\rangle \qquad \left[\begin{array}{c}\text{as } \mu(Y|x) = f(Y|x) \text{ wherever } \mu(x) > 0, \\ \text{and } \mu(X) \text{ minimizes } Inc_{\mathcal{m}_1}\end{array}\right] \end{aligned}$$

So $\mu(X,Y)$ witnesses the fact that $\langle\!\langle \mathcal{m}_2 \rangle\!\rangle \leq Inc_{\mathcal{m}_2}(\mu(X,Y)) = \langle\!\langle \mathcal{m}_1 \rangle\!\rangle$. Furthermore, every joint distribution $\nu(X,Y)$ must have at least this incompatibility, as it must have some marginal $\nu(X)$, which, even by itself, already gives rise to incompatibility of magnitude $Inc_{\mathcal{m}_1}(\nu(X)) \geq Inc_{\mathcal{m}_1}(\mu(X)) = \langle\!\langle \mathcal{m}_1 \rangle\!\rangle$. And since this is true for all $\nu(X,Y)$, we have that $\langle\!\langle \mathcal{m}_2 \rangle\!\rangle \geq \langle\!\langle \mathcal{m}_1 \rangle\!\rangle$. So $\langle\!\langle \mathcal{m}_2 \rangle\!\rangle = \langle\!\langle \mathcal{m}_1 \rangle\!\rangle$.

2. The equals sign in $\mathcal{m}_3$ may be equivalently interpreted as a cpd $eq(X_1|X_2) := x_2 \mapsto \delta_{x_2}(X_1)$, a cpd $eq'(X_2|X_1) := x_1 \mapsto \delta_{x_1}(X_2)$, or both at once; in each case, the effect is that a joint distribution $\mu$ with support on an outcome for which $X_1 \neq X_2$ gets an infinite penalty, so a minimizer $\mu(X_1, X_2, Y)$ of $Inc\mathcal{m}_3$ must be isomorphic to a distribution $\mu'(X,Y)$.

   Furthermore, it is easy to verify that $Inc_{\mathcal{m}_2}(\mu'(X,Y)) = Inc_{\mathcal{m}_3}(\mu(X,X,Y))$. More formally, we have:

$$\langle\!\langle \mathcal{m}_3 \rangle\!\rangle = \inf_{\mu(X_1,X_2,Y)} \mathop{\mathbb{E}}_{\mu} \left[\beta \log \frac{\mu(X_1)}{p(X_1)} + \zeta \log \frac{\mu(X_2)}{q(X_2)} + \beta \log \frac{\mu(Y|X_1)}{f(Y|X_1)} + \zeta \log \frac{\mu(Y|X_2)}{f(Y|X_2)} + \log \frac{\mu(X_1|X_2)}{eq(X_1,X_2)}\right]$$

   but if $X_1$ always equals $X_2$ (which we call simply $X$), as it must for the optimal $\mu$, this becomes

$$\begin{aligned} &= \inf_{\mu(X_1=X_2=X,Y)} \mathop{\mathbb{E}}_{\mu} \left[\beta \log \frac{\mu(X)}{p(X)} + \zeta \log \frac{\mu(X)}{q(X)} + \beta \log \frac{\mu(Y|X)}{f(Y|X)} + \zeta \log \frac{\mu(Y|X)}{f(Y|X)}\right] \\ &= \inf_{\mu(X,Y)} \mathop{\mathbb{E}}_{\mu} \left[\beta \log \frac{\mu(X)}{p(X)} + \zeta \log \frac{\mu(X)}{q(X)} + (\beta{+}\zeta) \log \frac{\mu(Y|X)}{f(Y|X)}\right] \\ &= \inf_{\mu(X,Y)} Inc_{\mathcal{m}_2}(\mu) \\ &= \langle\!\langle \mathcal{m}_2 \rangle\!\rangle. \end{aligned}$$

3. Eliminating the edge or edges enforcing the equality ($X_1 = X_2$) cannot increase inconsistency, by Lemma 1.

4. Although this final step of composing the edges with shared confidences looks intuitively like it should be true (and it is!), its proof may not be obvious. We now provide a rigorous proof of this equality.

   To ameliorate subscript pains, we henceforth write $X$ for $X_1$, and $Z$ for $X_2$. We now compute:

$$\begin{aligned} \langle\!\langle \mathcal{m}_4 \rangle\!\rangle &= \inf_{\mu(X,Z,Y)} \mathop{\mathbb{E}}_{\mu} \left[\beta \log \frac{\mu(X)\,\mu(Y|X)}{p(X)\,f(Y|X)} + \zeta \log \frac{\mu(Z)\,\mu(Y|Z)}{q(Z)\,f(Y|Z)}\right] \\ &= \inf_{\mu(X,Z,Y)} \mathop{\mathbb{E}}_{\mu} \left[\beta \log \frac{\mu(Y)\,\mu(X|Y)}{p(X)\,f(Y|X)} + \zeta \log \frac{\mu(Y)\,\mu(Z|Y)}{q(Z)\,f(Y|Z)}\right] \qquad \text{[apply Bayes Rule in numerators]} \end{aligned}$$

   By the chain rule, every distribution $\mu(X,Z,Y)$ may be specified as $\mu(Y)\mu(X|Y)\mu(Z|X,Y)$, so we can rewrite the formula above as

$$\langle\!\langle \mathcal{m}_4 \rangle\!\rangle = \inf_{\mu(Y)} \inf_{\mu(X|Y)} \inf_{\mu(Z|Y,X)} \mathop{\mathbb{E}}_{y\sim\mu(Y)} \mathop{\mathbb{E}}_{x\sim\mu(X|y)} \mathop{\mathbb{E}}_{z\sim\mu(Z|y,x)} \left[\beta \log \frac{\mu(y)\,\mu(x\,|\,y)}{p(x)\,f(y\,|\,x)} + \zeta \log \frac{\mu(y)\,\mu(z\,|\,y)}{q(z)\,f(y\,|\,z)}\right],$$

   where $\mu(Z|Y)$ is the defined in terms of the primitives $\mu(X|Y)$ and $\mu(Z|X,Y)$ as $\mu(Z|Y) := y \mapsto \mathbb{E}_{x\sim\mu(X|y)}\, \mu(Z|y,x)$, and is a valid cpd, since it is a mixture distribution. Since the first term (with $\beta$) does not depend on $z$, we can take it out of the expectation, so

$$\langle\!\langle \mathcal{m}_4 \rangle\!\rangle = \inf_{\mu(Y)} \inf_{\mu(X|Y)} \inf_{\mu(Z|Y,X)} \mathop{\mathbb{E}}_{y\sim\mu(Y)} \mathop{\mathbb{E}}_{x\sim\mu(X|y)} \left[\beta \log \frac{\mu(y)\,\mu(x\,|\,y)}{p(x)\,f(y\,|\,x)} + \zeta \mathop{\mathbb{E}}_{z\sim\mu(Z|y,x)} \left[\log \frac{\mu(y)\,\mu(z\,|\,y)}{q(z)\,f(y\,|\,z)}\right]\right];$$

we can split up $\mathbb{E}_{\mu(X|y)}$ by linearity of expectation, to get

$$\langle\!\langle m_4 \rangle\!\rangle = \inf_{\mu(Y)} \inf_{\mu(X|Y)} \inf_{\mu(Z|Y,X)} \mathop{\mathbb{E}}_{y\sim\mu(Y)} \left[ \beta \mathop{\mathbb{E}}_{x\sim\mu(X|y)} \left[ \log \frac{\mu(y)\,\mu(x\,|\,y)}{p(x)\,f(y\,|\,x)} \right] + \zeta \mathop{\mathbb{E}}_{\substack{x\sim\mu(X|y)\\ z\sim\mu(Z|y,x)}} \left[ \log \frac{\mu(y)\,\mu(z\,|\,y)}{q(z)\,f(y\,|\,z)} \right] \right]$$

Note that the quantity inside the second expectation does not depend on $x$. Therefore, the second expectation is just an explicit way of sampling $z$ from the mixture distribution $\mathbb{E}_{x\sim\mu(X|y)}\,\mu(Z|x,y)$, which is the definition of $\mu(Z|y)$. Once we make this replacement, it becomes clear that the only feature of $\mu(Z|Y,X)$ that matters is the mixture $\mu(Z|Y)$. Simplifying the second expectation in this way, and replacing the infemum over $\mu(Z|X,Y)$ with one over $\mu(Z|Y)$ yields:

$$\langle\!\langle m_4 \rangle\!\rangle = \inf_{\mu(Y)} \inf_{\mu(X|Y)} \inf_{\mu(Z|Y)} \mathop{\mathbb{E}}_{y\sim\mu(Y)} \left[ \beta \mathop{\mathbb{E}}_{x\sim\mu(X|y)} \left[ \log \frac{\mu(y)\,\mu(x\,|\,y)}{p(x)\,f(y\,|\,x)} \right] + \zeta \mathop{\mathbb{E}}_{z\sim\mu(Z|y)} \left[ \log \frac{\mu(y)\,\mu(z\,|\,y)}{q(z)\,f(y\,|\,z)} \right] \right]$$

Now, a cpd $\mu(X|Y)$ is just[7] a (possibly different) distribution $\nu_y(X)$ for every value of $Y$. Observe that, inside the expectation over $\mu(Y)$, the cpds $\mu(X|Y)$ and $\mu(Z|Y)$ are used only for the *present* value of $y$, and do not reference, say, $\mu(X|y')$ for $y' \neq y$. Because there is no interaction between the choice of cpd $\mu(X|y)$ and $\mu(X|y')$, it is not necessary to jointly optimize over entire cpds $\mu(X|Y)$ all at once. Rather, it is equivalent to to take the infemum over $\nu(X)$, separately for each $y$. Symmetrically, we may as well take the infemum over $\lambda(Z)$ separately for each $y$, rather than jointly finding the optimal $\mu(Z|Y)$ all at once. Operationallly, this means we can pull the infema inside the expectation over $Y$. And since the first term doesn't depend on $Z$ and the second doesn't depend on $X$, we get:

$$\langle\!\langle m_4 \rangle\!\rangle = \inf_{\mu(Y)} \mathop{\mathbb{E}}_{y\sim\mu(Y)} \left[ \inf_{\nu(X)} \beta \mathop{\mathbb{E}}_{\nu(X)} \left[ \log \frac{\mu(y)\,\nu(X)}{p(X)\,f(y\,|X)} \right] + \inf_{\lambda(Z)} \zeta \mathop{\mathbb{E}}_{\lambda(Z)} \left[ \log \frac{\mu(y)\,\lambda(Z)}{q(Z)\,f(y\,|Z)} \right] \right]$$

Next, we pull the same trick we've used over and over: find constants so that we can regard the dependence as a relative entropy with respect to the quantity being optimized. Grouping the quantities apart from $\nu(X)$ on the left term and normalizing them (and analogously for $\lambda(Z)$ on the right), we find that

$$\langle\!\langle m_4 \rangle\!\rangle = \inf_{\mu(Y)} \mathop{\mathbb{E}}_{y\sim\mu(Y)} \left[ \begin{array}{l} \beta \inf_{\nu(X)} \boldsymbol{D}\Big(\nu(X) \,\Big\|\, \frac{1}{C_1(y)} p(X) \frac{f(y|X)}{\mu(y)}\Big) - \beta \log C_1(y) \\ +\zeta \inf_{\lambda(Z)} \boldsymbol{D}\Big(\lambda(Z) \,\Big\|\, \frac{1}{C_2(y)} q(Z) \frac{f(y|Z)}{\mu(y)}\Big) - \zeta \log C_2(y) \end{array} \right],$$

where

$$C_1(y) = \sum_x p(x) \frac{f(y|x)}{\mu(y)} = \frac{1}{\mu(y)} \mathop{\mathbb{E}}_{p(X)} f(y|X) \qquad \text{and} \qquad C_2(y) = \sum_z q(z) \frac{f(y|z)}{\mu(y)} = \frac{1}{\mu(y)} \mathop{\mathbb{E}}_{q(Z)} f(y|Z)$$

are the constants required to normalize the distributions. Both relative entropies are minimized when their arguments match, at which point they contribute zero, so we have

$$\begin{aligned} \langle\!\langle m_4 \rangle\!\rangle &= \inf_{\mu(Y)} \mathop{\mathbb{E}}_{y\sim\mu(Y)} \left[ \beta \log \frac{1}{C_1(y)} + \zeta \log \frac{1}{C_2(y)} \right] \\ &= \inf_{\mu(Y)} \mathop{\mathbb{E}}_{y\sim\mu(Y)} \left[ \beta \log \frac{\mu(y)}{\mathbb{E}_{p(X)}\, f(y|X)} + \zeta \log \frac{\mu(y)}{\mathbb{E}_{q(Z)}\, f(y|Z)} \right] \\ &= \inf_{\mu(Y)} \mathop{\mathbb{E}}_{\mu} \left[ \beta \boldsymbol{D}(\mu \parallel f \circ p) + \zeta \boldsymbol{D}(\mu \parallel f \circ q) \right] \\ &= \langle\!\langle m_5 \rangle\!\rangle. \end{aligned}$$

[7] modulo measurability concerns that do not affect the infemum; see Appendix A

### C.1.2 Details for Claims made in Section 8

First, the fact that

$$\mathcal{L}_1 = \lambda_{\mathsf{d}}\mathcal{L}_{\mathtt{dat}} + \lambda_{\mathsf{s}}\mathcal{L}_{\mathtt{sim}} =$$

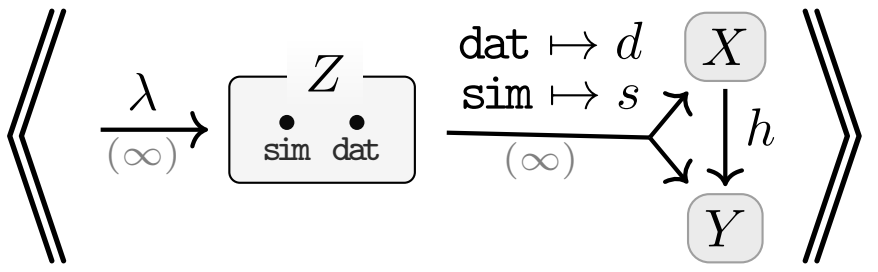

,

where $\lambda(Z = \mathtt{sim}) = \lambda_{\mathsf{s}}$ and $\lambda(Z = \mathtt{dat}) = \lambda_{\mathsf{d}}$ is immediate. The two cpds with infinite confidence ensure that the only joint distribution with a finite score is $\lambda_{\mathsf{s}} s + \lambda_{\mathsf{d}} d$, and the inconsistency with $h$ is its surprisal, so the inconsistency of this PDG is

$$\mathop{\mathbb{E}}_{\lambda_{\mathsf{s}} s + \lambda_{\mathsf{d}} d}\left[\log \frac{1}{h(Y|X)}\right] = -\lambda_{\mathsf{s}} \mathop{\mathbb{E}}_{s}[\log h(Y|X)] - \lambda_{\mathsf{d}}\, \mathbb{E}\, d[\log h(Y|X)] = \lambda_{\mathsf{d}}\mathcal{L}_{\mathtt{dat}} + \lambda_{\mathsf{s}}\mathcal{L}_{\mathtt{sim}} = \mathcal{L}_1, \quad \text{as promised.}$$

The second correspondence is the least straightforward. Let $C = \int sd$ be the normalization constant required to normalize the joint density $sd$. We claim that, for large fixed $\gamma$, we have

$$\mathcal{L}_2 \approx C \left\langle\!\!\left\langle \quad \right\rangle\!\!\right\rangle_\gamma + \mathit{const},$$

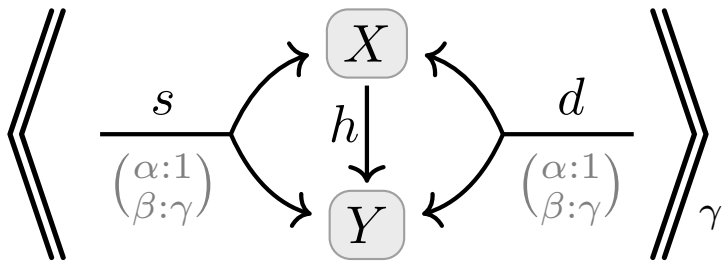

where *const* does not depend on $h$. To see this, let $\mathcal{M}_2$ be the PDG above, and compute

$$\begin{aligned}
\langle\!\langle \mathcal{M}_2 \rangle\!\rangle_\gamma &= \inf_{\mu(X,Y)} \mathop{\mathbb{E}}_{\mu}\Bigg[ \overbrace{\gamma \log \frac{\mu(XY)}{s(XY)}\frac{\mu(XY)}{d(XY)} + \log \frac{\mu(Y|X)}{h(Y|X)}}^{Inc(\mu)} + \overbrace{\gamma \log \frac{1}{s(XY)}\frac{1}{d(XY)} - \gamma \log \frac{1}{\mu(XY)}}^{IDef(\mu)} \Bigg] \\
&= \inf_{\mu(X,Y)} \mathop{\mathbb{E}}_{\mu}\left[ \gamma \log \frac{\mu(XY)}{s(XY)}\frac{\mu(XY)}{d(XY)}\frac{1}{\mu(XY)}\frac{1}{\mu(XY)}\frac{\mu(XY)}{1} + \log \frac{\mu(Y|X)}{h(Y|X)} \right] \\
&= \inf_{\mu(X,Y)} \mathop{\mathbb{E}}_{\mu}\left[ \gamma \log \frac{\mu(XY)}{s(XY)d(XY)} + \log \frac{\mu(Y|X)}{h(Y|X)} \right] \\
&= \inf_{\mu(X,Y)} \mathop{\mathbb{E}}_{\mu}\left[ \gamma \log \frac{\mu(XY)C}{s(XY)d(XY)} - \gamma \log C + \log \frac{\mu(Y|X)}{h(Y|X)} \right] \\
&= \inf_{\mu(X,Y)} \gamma \boldsymbol{D}\left(\mu \,\middle\|\, \frac{1}{C} sd\right) + \mathop{\mathbb{E}}_{\mu}\left[ \log \frac{\mu(Y|X)}{h(Y|X)} \right] - \gamma \log C
\end{aligned}$$

$\boldsymbol{D}$ is $(\gamma m)$-strongly convex in a region around its minimizer for some $m > 0$ that depends only on $s$ and $d$. Together with our assumption that $h$ is positive, we find that when $\gamma$ becomes large, the first term dominates, and the optimizing $\mu$ quickly approaches the normalized density $\nu := \frac{1}{C} sd$. Plugging in $\nu$, we find that the value of the infemum approaches

$$\begin{aligned}
\langle\!\langle \mathcal{M}_2 \rangle\!\rangle &\approx \mathop{\mathbb{E}}_{\nu}\left[ \log \frac{1}{h(Y|X)} \right] - H_\nu(Y|X) - \gamma \log C \\
&= \int_{XY} \frac{1}{C} \log \frac{1}{h(Y|X)} s(X,Y) d(X,Y) \quad - H_\nu(Y|X) - \gamma \log C \\
&= \frac{1}{C} \mathop{\mathbb{E}}_{s}\left[ d(X,Y) \log \frac{1}{h(Y|X)} \right] - H_\nu(Y|X) - \gamma \log C \\
&= \frac{1}{C} \mathcal{L}_2 - H_\nu(Y|X) - \gamma \log C,
\end{aligned}$$

$$\text{and therefore} \qquad \begin{aligned} \mathcal{L}_2 &= C \langle\!\langle \mathcal{M}_2 \rangle\!\rangle + C\, \mathrm{H}_\nu(Y|X) - \gamma\, C \log C \\ &= C \langle\!\langle \mathcal{M}_2 \rangle\!\rangle + \mathit{const}. \end{aligned}$$

Finally, we turn to

$$\mathcal{L}_3 :=$$ 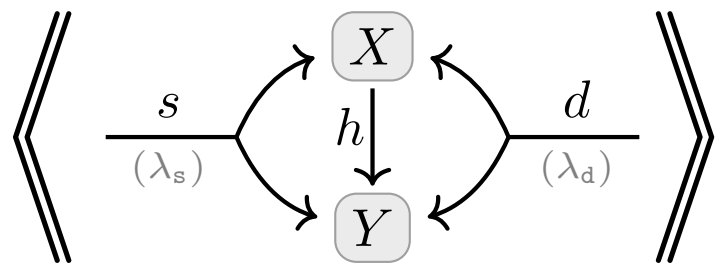

To see the why the optimal distribution $\mu^*(XY)$ is the $\lambda$-weighted geometric mean of $s$ and $d$, let us first consider the same PDG, except without $h$. From Lemma 10, we have this loss without $h$ in closed form, and from the proof of Lemma 10, we see that the optimizing distribution in this case is the $\lambda$-weighted geometric distribution $\mu^* \propto s(XY)^{\lambda_\mathrm{s}} d(XY)^{\lambda_\mathrm{d}}$. Now (Lemma 1), including $h$ cannot make the PDG any less inconsistent. In particular, by choosing

$$h^*(Y|X) := \mu^*(Y|X) \propto (Y|X)^{\lambda_\mathrm{s}} d(Y|X)^{\lambda_\mathrm{d}},$$

to be already compatible with this joint distribution, the inconsistency does not change, while choosing a different $h$ would cause the inconsistency to increase. Thus, the optimal classifier $h^*$ by this metric is indeed as we claim. Finally, it is easy to see that this loss is calibrated: if $s = d$, then the optimal joint distribution is equal to $s$ and to $d$, and the optimal classifier is $h(Y|X) = s(Y|X) = d(Y|X)$. So $\mathcal{L}_3$ is calibrated.

### C.1.3 Details for Claims made in Section 9

**Distortion Due to Inconsistency.** In the footnote on Page 9, we claimed that if the model confidence $\beta_p$ were 1 rather than $\infty$, we would have obtained an incconsistency of $-\log \mathbb{E}_{x\sim p} \exp(-c(x))$, and that the optimal distribution would not have been $p(X)$.

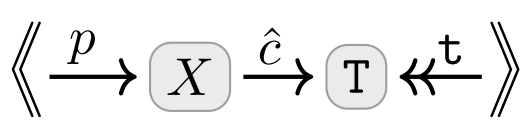

$$\begin{aligned} &= \inf_{\mu(X)} \mathop{\mathbb{E}}_{x\sim\mu} \left[ \log \frac{\mu(x)}{p(x)} + \log \frac{\mu(\mathtt{t}\,|\,x)}{\hat{c}(\mathtt{t}\,|\,x)} \right] \\ &= \inf_{\mu(X)} \mathop{\mathbb{E}}_{x\sim\mu} \left[ \log \frac{\mu(x)}{p(x)} + \log \frac{1}{\hat{c}(\mathtt{t}\,|\,x)} \right] \\ &= \inf_{\mu(X)} \mathop{\mathbb{E}}_{x\sim\mu} \left[ \log \frac{\mu(x)}{p(x)\exp(-c(x))} \cdot \frac{Z}{Z} \right] \end{aligned}$$

where $Z = \sum_x p(x)\exp(-c(x)) = \mathbb{E}_p \exp(-c(X))$ is the constant required to normalize the distribution

$$\begin{aligned} &= \inf_{\mu(X)} \boldsymbol{D}\left( \mu \;\middle\|\; \frac{1}{Z} p(X)\exp(-c(X)) \right) - \log Z \\ &= -\log Z \\ &= -\log \mathop{\mathbb{E}}_{x\sim p} \exp(-c(x)) \end{aligned}$$

as promised. Note also that in the proof, we showed that the optimal distribution is proportional to $p(X)\exp(-c(X))$ which means that it equals $p(X)$ if and only if $c(X)$ is constant in $X$.

**Enforcing the Qualitative Picture.** We also claimed without careful proof in Section 9 that, if $\alpha_h = \alpha_{\Pr_\mathcal{D}} = 1$, then

$$\lim_{\gamma\to\infty}$$ 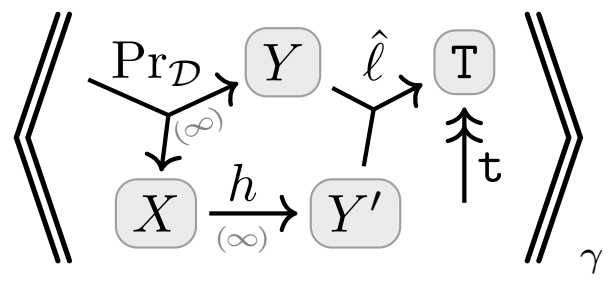

$$= \mathop{\mathbb{E}}_{\substack{(x,y)\sim\Pr_\mathcal{D} \\ y'\sim p(Y'\,|\,x)}} \left[ \ell(y,y') \right]$$

Why is this? For such a setting of $\alpha$, which intuitively articulates a causal picture where $X, Y$ is generated from $\Pr_\mathcal{D}$, and $Y'$ generated by $h(Y'|X)$, the information deficiency $\mathit{IDef}_\mathcal{S}(\mu(X,Y,Y'))$ of a distribution $\mu$ is

$$\begin{aligned} \mathit{IDef}_\mathcal{S}(\mu(X,Y,Y')) &= -\mathrm{H}_\mu(X,Y,Y') + \mathrm{H}(X,Y) + \mathrm{H}(Y'|X) \\ &= \mathrm{H}_\mu(Y'|X) - \mathrm{H}_\mu(Y'|X,Y) \\ &= \mathrm{I}_\mu(Y;Y'|X). \end{aligned}$$

Both equalities of the derivation above standard information theoretic identities (See, for instance, MacKay 2003), and the final quantity $\mathrm{I}_\mu(Y;Y'|X)$ is the *conditional mutual information* between $Y$ and $Y'$ given $X$, and is a non-negative number that equals zero if and only if $Y$ and $Y'$ are conditionally independent given $X$.

As a result, as $\gamma \to \infty$ any distribution that for which $Y'$ and $Y$ are not independent given $X$ will incur infinite cost. Since the confidences in $h$ and $\Pr_\mathcal{D}$ are also infinite, so will a violation of either cpd. There is only one distribution that has both cpds and also this independence; that distribution is $\mu(X,Y,Y') := \Pr_\mathcal{D}(X,Y)h(Y'|X)$. Now the argument of Proposition 15 applies: all other cpds must be matched, and the inconsistency is the expected incompatibility of $\hat{l}$, which equals

$$\mathop{\mathbb{E}}_{\substack{(x,y)\sim\Pr_\mathcal{D}\\ y'\sim p(Y'|x)}} \log \frac{1}{\hat{\ell}(\mathtt{t}\,|y,y')} = \mathop{\mathbb{E}}_{\substack{(x,y)\sim\Pr_\mathcal{D}\\ y'\sim p(Y'|x)}} \log \frac{1}{\exp(-\ell(y,y'))} = \mathop{\mathbb{E}}_{\substack{(x,y)\sim\Pr_\mathcal{D}\\ y'\sim p(Y'|x)}} \Big[\log \exp(\ell(y,y'))\Big] = \mathop{\mathbb{E}}_{\substack{(x,y)\sim\Pr_\mathcal{D}\\ y'\sim p(Y'|x)}} \Big[\ell(y,y')\Big] = \mathcal{L}.$$

# D More Notes

## D.1 Maximum A Posteriori and Priors

The usual telling of the correspondence between regularizers and priors is something like the following. Suppose you have a parameterized family of distributions $\Pr(X|\Theta)$ and have observed evidence $X$, but do not know the parameter $\Theta$. The maximum-likelihood estimate of $\Theta$ is then

$$\theta^{\mathrm{MLE}}(X) := \arg\max_{\theta\in\Theta} \Pr(X|\theta) = \arg\max_{\theta\in\Theta} \log \Pr(X|\theta).$$

The logarithm is a monotonic transformation, so it does not change the argmax, but it has nicer properties, so that function is generally used instead. (Many of the loss functions in main body of the paper are log-likelihoods also.)

In some sense, better than estimating the maximum likelihood, is to perform a Bayesian update with the new information, to get a *distribution* over $\Theta$. If that's too expensive, we could simply take the estimate with the highest posterior probability, which is called the Maximum A Posteriori (MAP) estimate. For any given $\theta$, the Bayesian reading of Bayes rule states that

$$\text{posterior } \Pr(\Theta|X) = \frac{\text{likelihood } \Pr(X|\Theta) \cdot \text{prior } \Pr(\Theta)}{\text{evidence } \Pr(X) = \sum_{\theta'} \Pr(X|\theta')\Pr(\theta')}.$$

So taking a logarithm,

$$\text{log-posterior } \log \Pr(\Theta|X) = \text{log-likelihood } \log \Pr(X|\Theta) \;+\; \text{log-prior } \log \Pr(\Theta) - \text{log-evidence } \log \Pr(X).$$

The final term does not depend on $\theta$, so it is not relevant for finding the optimal $\theta$ by this metric. Swapping the signs so that we are taking a minimum rather than a maximum, the MAP estimate is then given by

$$\theta^{\mathrm{MAP}}(X) := \arg\min_{\theta\in\Theta} \left\{ \log \frac{1}{\Pr(X|\theta)} + \log \frac{1}{\Pr(\theta)} \right\}.$$

Note that if negative log likelihood (or surprisal, $-\log \Pr(X|\theta)$) was our original loss function, we have now added an arbitrary extra term, as a function of $\Theta$, to our loss function. It is in this sense that priors classically correspond to regularizers.